\documentclass[10pt,twocolumn,letterpaper]{article}

\usepackage[pagenumbers]{cvpr} % To force page numbers, e.g. for an arXiv version
\usepackage{multirow}

\usepackage[dvipsnames]{xcolor}

\definecolor{cvprblue}{rgb}{0.21,0.49,0.74}
\usepackage[pagebackref,breaklinks,colorlinks,citecolor=cvprblue]{hyperref}

\def\paperID{9261} % *** Enter the Paper ID here
\def\confName{CVPR}
\def\confYear{2024}

\title{Data-free Multi-label Image Recognition via LLM-powered Prompt Tuning}

\author{Shuo Yang\textsuperscript{\rm 1,2} \quad Zirui Shang\textsuperscript{\rm 2} \quad Yongqi Wang\textsuperscript{\rm 2} \\ 
Derong Deng\textsuperscript{\rm 1} \quad Hongwei Chen\textsuperscript{\rm 1} \quad Qiyuan Cheng\textsuperscript{\rm 1} \quad Xinxiao Wu\textsuperscript{\rm 1,2}\thanks{Corresponding author.}\\
\textsuperscript{\rm 1}Guangdong Laboratory of Machine Perception and Intelligent Computing\\Shenzhen MSU-BIT University, China\\
\textsuperscript{\rm 2}Beijing Key Laboratory of Intelligent Information Technology\\School of Computer Science \& Technology, Beijing Institute of Technology, China\\
\tt\small \{shuoyang,shangzirui,wuxinxiao\}@bit.edu.cn \\ 
\tt\small \{1285441164yq, derongdeng.dero, chwr0001\}@gmail.com \quad chengqiyuan@smbu.edu.cn
}
\begin{document}
\maketitle
\begin{abstract}{}
% 可以参考 Visual Programming这篇论文的abstract写作方式。
% 1. 框架提出-->怎么解决使用LLM解决data free，以及独特的使用方式-->prompting-->实验结果
This paper proposes  a  novel  framework for multi-label image recognition without any training data, called data-free framework, which uses knowledge of pre-trained Large Language Model (LLM) to learn prompts to adapt pre-trained  Vision-Language  Model (VLM) like 
CLIP to  multi-label classification. % Leveraging a pre-trained large language model (LLM), we extract information about the characteristics of objects and their relationships by designing organized and structured purposeful questions. This information aids in prompt-tuning CLIP to enhance its multi-label image recognition capabilities. 
% Through interacting with LLM by well-designed questions, we obtain description/knowledge about characteristics of objects, which provides a comprehensive understanding of objects and serves as valuable data for prompts learning.
Through asking LLM by well-designed questions, we acquire comprehensive knowledge about characteristics and contexts of objects, which provides valuable text descriptions for learning prompts.
% We demonstrate LLM can be an alternative way to provide valuable information for model optimization.
% Specifically, through questioning LLM, we first extract the attributes and corresponding descriptions of the objects, and then imagine different scenarios that different objects present to obtain relationship information.
% This knowledge is then employed to establish a hierarchical prompt tuning paradigm, sharing tokens of prompts among different objects based on their co-occurrence and relationships.
Then we propose a hierarchical prompt learning method by taking the multi-label dependency  into consideration, wherein a subset of category-specific prompt tokens are shared when the corresponding objects exhibit similar attributes or are more likely to co-occur.
% To take the label-dependency of multi-label image recognition into consideration, we introduce a hierarchical prompt learning method wherein a subset of class-specific prompt tokens is shared when the corresponding objects exhibit similar attributes or are more likely to co-occur in certain scenarios.
% Benefiting from the remarkable alignment between visual and linguistic semantics of CLIP, the learned hierarchical prompts are  applied to perform image classification during inference. 
Benefiting from the remarkable alignment between visual and linguistic semantics of CLIP, the hierarchical prompts learned from text descriptions are applied to perform classification of images during inference. 
%Our framework presents a potential direction that recognizes novel objects in images by pre-trained models and 
Our framework presents a new way to explore the synergies between multiple pre-trained models for novel category recognition.
Extensive experiments on three public datasets (MS-COCO, VOC2007, and NUS-WIDE) demonstrate that our method achieves better results than the state-of-the-art methods, especially outperforming the zero-shot multi-label  recognition methods by 4.7\% in mAP on MS-COCO.

\end{abstract}

\section{Introduction}
\label{sec:intro}

\begin{figure}
    \centering
    \includegraphics[width=\linewidth]{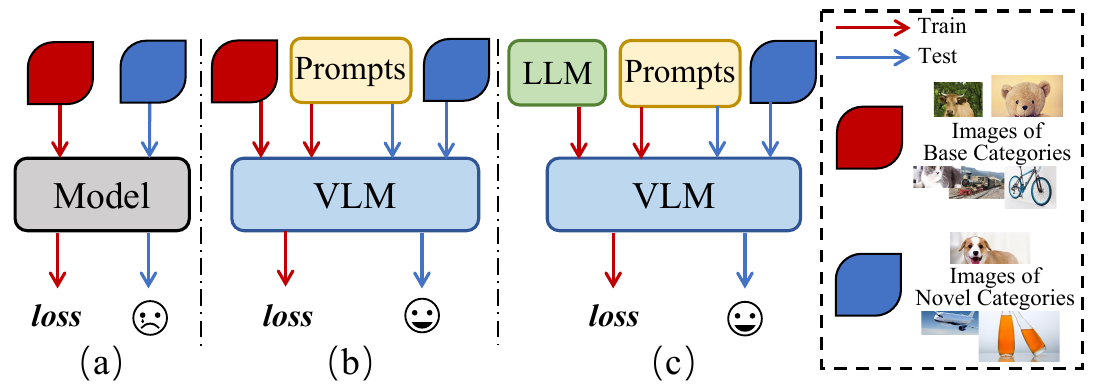}
    \caption{Illustration of different ways to handle novel categories. (a) Traditional methods train on base categories but fail on novel categories. (b) Recent prompting methods successfully adapt VLM to novel categories but need annotated data for prompt tuning. (c) Our data-free framework only performs prompt tuning to adapt VLM to novel categories by LLM.}
    \label{fig:intro}
\end{figure}

% multi-label --> pre-trained model的介入-->text as image -->our method 提出-->细节 -->优势，以及和其他使用LLM方法的对比 --> contribution
% which typically deals with images of a diverse array of objects and scenes. 
Multi-label image recognition aims to recognize all objects present in an image.
This task is challenging  due to the  emergence of novel objects and scenes~\cite{ben2021semantic} during inference in real-world scenarios, as shown in Figure~\ref{fig:intro}(a). 
% Efforts have been made to mitigate this problem by developing appropriate unsupervised or weakly supervised methods~\cite{xxx}. However, due to poor performance, there is still a long way to go before practical application.
% Early efforts have been made to alleviate this issue by formulating appropriate open-set problems~\cite{6365193}. 
% Nonetheless, within most current open-set frameworks, the prevalent approach when confronted with newly encountered classes is to designate them as an ``unknown" category~\cite {yoshihashi2019classification}, thus limiting their practical applicability.
Recent large-scale pre-trained Vision-Language Models (VLMs) like CLIP~\cite{radford2021learning} %and ChatGPT~\cite{bubeck2023sparks},  
spawn the training-free zero-shot methods~\cite{guo2023calip}, which can handle new categories by calculating similarities between images and texts in a well-aligned embedding space. To further effectively adapt VLMs to enhance the performance of novel categories, several methods have been proposed to learn adapter~\cite{abdelfattah2023cdul} or prompts~\cite{sun2022dualcoop} using sufficient  annotated images, as shown in Figure~\ref{fig:intro}(b).
However, the performance of these prompt learning methods may be limited when it is infeasible to collect sufficient fully annotated images. % can not be accessed for these methods, limiting their practical applicability. 

To address this issue, Sun \textit{et al.}~\cite{sun2022dualcoop} propose dual context optimization to quickly adapt CLIP to multi-label recognition using partially labeled images, where only a few categories for each training image are annotated, significantly reducing the annotation burden. Guo \textit{et al.}~\cite{guo2023texts} propose texts as images in prompt tuning to adapt CLIP, where the text  descriptions are human-written image captions from existing datasets and serve as alternatives to  images. This method presents a more practical and efficient way for prompting  as text descriptions are more easily accessible  than images.

% This method collects text captions from existing datasets and treats them as alternatives of images, eliminating the need for actual images. Though texts can be acquired through the internet, which is easier than gathering images, this process is time-consuming and requires additional post-processing. 
% It is challenging to conjure an image of an entirely unfamiliar object, but we can effectively identify the object in a picture if provided with a linguistic description. This observation inspires us to glean knowledge from language and apply it to images.
% Language knowledge can be acquired through internet search engines. While relatively easier than gathering images, this process is time-consuming.
% we introduce a novel framework: initially learning knowledge through the language modality and subsequently transferring the corresponding knowledge to the visual domain. 
% Language knowledge can be acquired through expert input or internet search engines. While relatively easier than gathering images, this process can be time-consuming and resource-intensive. Is there a more efficient alternative approach? 
%In this paper, we propose a data-free framework that relies neither on images nor texts but leverages an additional Large Language Model (LLM), as shown in Figure~\ref{fig:intro} (c). 
%We adapt CLIP for multi-label image recognition through prompt tuning based on the knowledge extracted by directly asking LLM well-designed questions.
In this paper, we propose a data-free framework for multi-label image recognition without any data for training. It leverages knowledge of objects from pre-trained Large Language Model (LLM)  to adapt CLIP to multi-label classification by textual prompt tuning,  as shown in Figure~\ref{fig:intro}(c).
Specifically, we propose to collect  comprehensive information of objects by designing different types of  questions posed to LLM. 
Starting with asking LLM category-agnostic questions like \textit{[object lists], please summarize 90 attributes that may be common to the above 80 words} to acquire common attributes, such as shape, color, and material, shared by all categories, similarly we then acquire  particular attributes for each category by category-specific questions like \textit{please summarize 30 attributes of [object]}.  
Finally, we acquire text descriptions of the attributes by category-description questions like \textit{please help me generate 100 different sentences about [category] from the angle of the [attribute]}. 
% What's more, real-world scenes relevant to multiple objects provide rich context information, which can be obtained by asking LLM questions like ``generate ten sentences to describe different scenes involving [category1] and [category2]", namely scene-related knowledge.
%The resulting object attribute knowledge and scene-related knowledge, along with the object name, serve as our training dataset.
Moreover, we design scene-related questions like \textit{generate ten sentences to describe different scenes involving [category1] and [category2]} to acquire text descriptions of contextual relationships between multiple object categories in real-world scenes, namely relationship knowledge. 
Along with category labels, the acquired text descriptions of attribute and relationship knowledge  from LLM are used as images for prompt tuning CLIP to multi-label recognition. 
To incorporate the  relationship information between multiple objects into prompt learning to further improve  the performance, 
%Motivated by the fact that incorporating relationships between  objects enhances the multi-label recognition performance  significantly~\cite{lee2018multi}, 
we propose a hierarchical prompt learning method, which categorizes the  prompt tokens into three types: (1) shared tokens shared by all object categories; (2) partial-shared tokens  shared by the object categories of the same subgroups with co-occurrence relationship or similar attributes; (3) category-specific tokens specific to each individual object category. 
Through designing these hierarchical tokens, we learn prompts that absorb both task-specific knowledge and object-specific knowledge, as well as the relationship knowledge between objects.
% Thanks to the excellent alignment of visual and linguistic semantics of CLIP, we learn hierarchical prompts in the language aspect and then apply them to the image inputs during inference. 
Benefiting from the remarkable alignment between visual and linguistic semantics of CLIP, the hierarchical prompts learned from text description are applied to perform classification of images during inference.
%By doing so, we improve the performance of multi-label image recognition by a large margin compared to the simple zero-shot approaches. 

In summary, the  contributions of our work are three-fold: 
\begin{itemize}
\item  We propose a data-free framework for multi-label image recognition without any training data, which leverages rich knowledge in LLM to prompt tune CLIP. Our framework introduces a promising avenue for handling new objects in visual recognition, relying solely on pre-trained models, and also paves an effective way to  explore the synergies between multiple pre-trained models.
% (2) we design several questions to acquire knowledge of given object categories from LLMs, both by their own characteristics and their co-occurrence and relationships.
\item  We propose a hierarchical prompt learning method to  adapt CLIP by using the acquired knowledge of objects from LLM. It incorporates relationships between different categories into learnable prompts, thus further improving the multi-label recognition performance.
\item  We propose to collect comprehensive information about object attributes and  relationships from LLM by designing different types of questions.
%Our framework introduces a promising avenue for recognizing new objects in images, relying solely on pre-trained models. Moreover, it paves a new way for further exploration of the synergies between large pre-trained models.
\end{itemize}
\section{Related Work}
\label{sec:related}
\subsection{Multi-Label Image Recognition}
Early multi-label image recognition methods~\cite{wei2015hcp} naively treat this task as a multiple independent binary classification problem, which trains a binary classifier for each category~\cite{liu2015optimality,misra2016seeing}. 
However, these methods do not consider correlations among labels, and recent works have focused on incorporating semantic dependencies among labels via graph modeling~\cite{chua2009nus,lee2018multi,chen2019multi,chen2019learning,wang2020multi} or sequential perdition~\cite{wang2016cnn,liu2017semantic,wang2017multi,he2018reinforced,zhang2018multilabel,yazici2020orderless}. 
There are also works focused on the attention mechanism~\cite{sarafianos2018deep,zhu2017learning,gao2020multi,ye2020attention} or loss functions~\cite{gong2013deep,li2017improving,ridnik2021asymmetric}. 

% Despite notable progress has been made, it's worth noting that these methods, while effective, require a large-scale annotated dataset for model training. 
% This limits their application in data-limited or label-limited regimes, where the proposed models and specifically designed loss functions may fail. 
% Thus, a bunch of method struggle to few-shot/zero-shot~\cite{lee2018multi,alfassy2019laso,ji2020deep,simon2022meta,ji2020deep} and partial-labeled~\cite{durand2019learning,abdelfattah2022plmcl,chen2022structured,pu2022semantic,zhang2023learning} multi-image recognition. 

Despite  remarkable progress that has been made, these methods heavily on large-scale annotated images for training, which limits their capabilities in data-limited or label-limited scenarios. %, where the proposed models and specifically designed loss functions may fail.
In recent years, several methods have emerged to address the  few-shot/zero-shot multi-label image recognition ~\cite{lee2018multi,alfassy2019laso,ji2020deep,simon2022meta,huynh2020shared,ben2021semantic} and partial-labeled image recognition~\cite{durand2019learning,abdelfattah2022plmcl,chen2022structured,pu2022semantic,zhang2023learning}. 
By exploring the synergies between LLM and VLM,  we take a significant step forward in multi-label image recognition by introducing a data-free framework where no training data is provided. 

% This innovative approach offers a novel and complementary perspective for addressing the challenges of low-resource multi-label image recognition.

\begin{figure*}
    \centering
    \includegraphics[width=\textwidth]{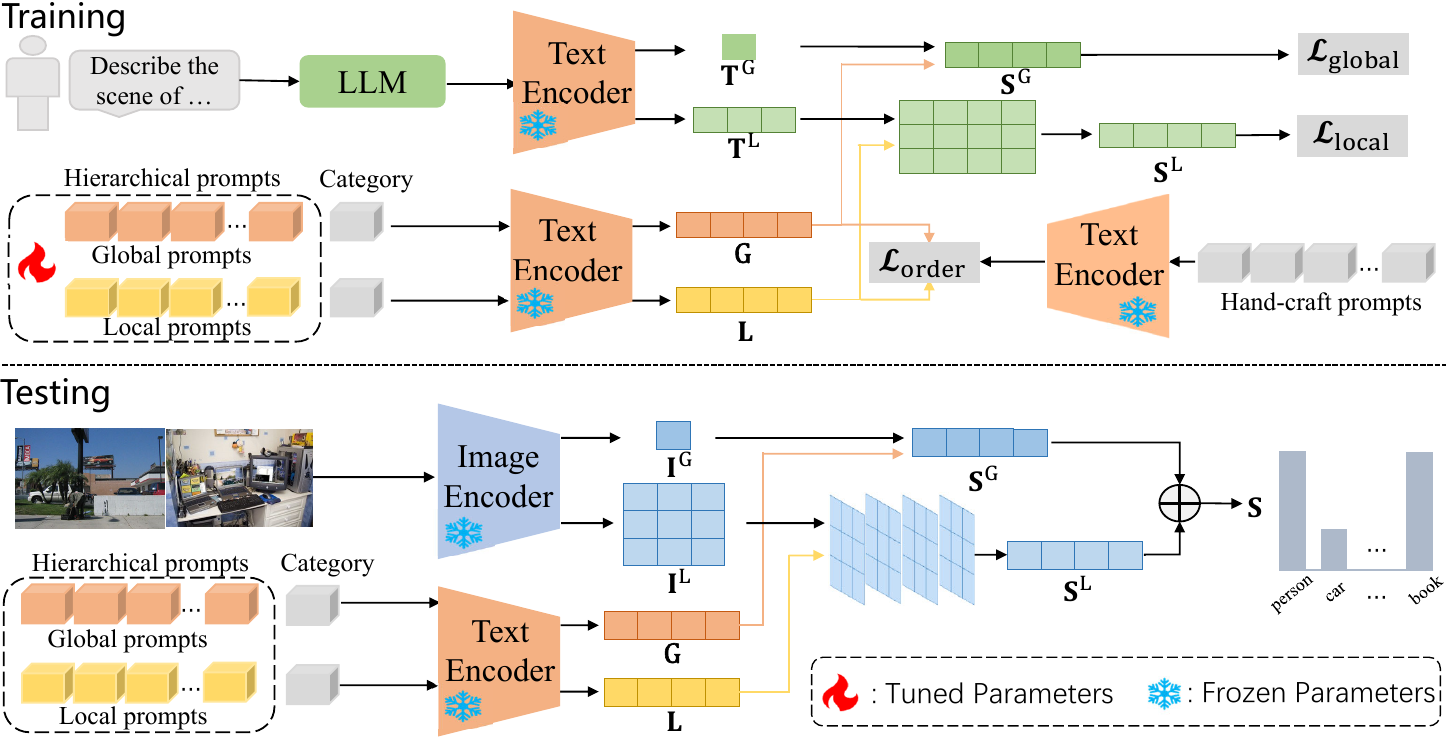}
    \caption{Overview of our framework.}
    \label{fig:framework}
\end{figure*}

% \begin{figure*}[t]
%     \centering
%     \begin{tabular}{cc}
%         \includegraphics[width=0.8\linewidth]{figs/pipeline_train.pdf} \\  \includegraphics[width=0.8\linewidth]{figs/pipeline_test.pdf}
%     \end{tabular}
%     \caption{Overview of the proposed framework.}
%     \label{fig:framework}
% \end{figure*}

\subsection{Adapting CLIP to Visual Tasks}
% Large pre-trained models boost most AI tasks, from natural language processing~\cite{xxx} to visual understanding~\cite{xxx}.  
Vision-Language Models (VLMs) have demonstrated impressive capabilities on learning generic representations, such as CLIP~\cite{radford2021learning}.
% BLIP~\cite{li2022blip}.
In order to adapt  VLMs to  specific downstream tasks, many prompt tuning methods~\cite{yao2021cpt,jia2022visual,zhou2022learning,zhou2022conditional,sohn2023visual,zhu2023prompt,singha2023ad} have been proposed to learn task-specific prompts, which gains significant attention for both excellent performance and parameter-efficient characteristic. 
To further to bridge the domain gap between the data used to train VLMs and that of  specific tasks,  dedicated adapters~\cite{zhang2022tip,sung2022vl,upadhyay2023probvlm,gao2023clip,xu2023side,chen2023tem,abdelfattah2023cdul} have been designed and integrated into CLIP, % These adapters are tailored to specific domains, 
avoiding fine-tuning the entire model.
% Similarly, prompt-tuning methods~\cite{yao2021cpt,jia2022visual,zhou2022learning,zhou2022conditional,sohn2023visual,zhu2023prompt,singha2023ad} gain significant attention. These methods learn task-specific prompts to utilize the pre-trained model for downstream tasks and have gained prominence for their parameter-efficient characteristics.
% 

The works most relevant to our method are DualCoOp~\cite{sun2022dualcoop} and TaI-DPT~\cite{guo2023texts}. DualCoOp learns a pair of differentiable prompts %to provide
%positive and negative contexts for the target class 
using partial-annotated images, and  TaI-DPT uses image captioning collected from existing datasets as images to learn prompts. In contrast, our method inquires LLMs to acquire comprehensive knowledge of object categories as text description for prompt learning. Moreover, our method learns relationship-aware hierarchical prompts which are tailored to multi-label image recognition.
%Considering the relationships between categories, this paper based on the generated textual descriptions, which are responses of LLMs by asking well-designed questions about various aspects of the given objects. 

\subsection{LLM-enhanced Visual Understanding}
LLMs have been allied to help visual understanding tasks~\cite{yang2023llm,brown2020language,cai2023low} due to their ``emergent abilities" of learning how to answer such questions from the in-context examples~\cite{wei2022emergent}.  
Visual information is represented as text descriptions and then fed into the LLMs together with the target question and in-context examples to generate the desired results~\cite{yang2022empirical,gupta2023visual}. 
A recent study employs ChatGPT~\cite{bubeck2023sparks} to generate a comparison tree, which enhances CLIP's zero-shot performance on image classification. 
In this paper, we focus on using LLM to generate text descriptions to facilitate prompt tuning of CLIP to multi-label image recognition.

\section{Our Method}
\label{sec:methods}
\subsection{Overview}
We propose a data-free framework for multi-label
image recognition without any training data. A  large language model (ChatGLM) serves as a repository of encyclopedic knowledge, and we propose to acquire  comprehensive knowledge of  object categories by designing different types of questions posed to ChatGLM. This is motivated by the fact that we can effectively identify an object in a picture if provided with a linguistic description. Then a pre-trained vision-language model (CLIP) is prompt tuned using the acquired knowledge to enhance multi-label classification, based on the aligned visual and linguistic embedding space. We propose a hierarchical prompt learning method to incorporate relationships between objects into the learnable prompts. Figure~\ref{fig:framework} shows an overview of our framework.

Given an input image $\textbf{x}$, multi-label image recognition aims to identify all object categories in it, formulated as $\mathbf{S} = f_{\Phi, \Psi}(\mathbf{x})$, where $f_{\Phi, \Psi}$ denotes the recognition model,  $ \Phi$ denotes ChatGLM~\cite{du2022glm}, $\Psi$ denotes CLIP, including  a text encoding module $\Psi_t$ and an image encoding module $\Psi_i$, and $\mathbf{S} \in \mathbb{R}^{N}$ is the predicted probability scores for all $N$ categories $\mathbb{Y}=\{Y_1,\cdots,Y_N\}$.

\subsection{Knowledge Acquirement}
%For describing an object, we need critical information about its color, shape, texture, and so on. 
%To get such information, we ask well-designed questions to the LLM , which is known for its rich encyclopedic knowledge and serves as a chatbot responding to the input questions.

%来自chatGPT:
To describe an object, it is crucial to have detailed information about its color, shape, texture, and other attributes. To obtain this information, we engage with ChatGLM, a highly knowledgeable language model that functions as a chatbot and responds to carefully crafted questions. Its extensive encyclopedic knowledge allows us to extract the necessary details for adequate object description.

\noindent \textbf{Coarse Attribute Description.}
%To get diverse aspects of objects, we first obtain various attributes that can describe and discriminate different objects by asking ChatGLM ($\Phi$) well-designed questions:
To capture diverse aspects of objects, we begin by extracting common attributes shared by all categories using category-agnostic questions and particular attributions for individual categories 
using category-specific questions, formulated as %attributes that can effectively describe and differentiate between various objects.
\begin{equation} \label{eq:coarse}
   ( \mathbb{A}_c,\mathbb{A}_s) = \Phi(\Pi_1(\mathbb{Y})),
\end{equation}
where $\Pi_1(\cdot)$ denotes the common and category-specific questions like \textit{[object lists], please summarize 90 attributes that may be common to the above 80 words}. $\mathbb{A}_c = \{a_1,\cdots,a_{n_1}\}$ denotes  $n_1$ common attributes of all categories.  
$\mathbb{A}_s=\{\mathbb{A}_{s,1},\cdots,\mathbb{A}_{s,N}\}$ denotes the category-specific attribute sets, where $\mathbb{A}_{s,i}=\{a_{i,1},\cdots, a_{i,n_2}\}$ denotes $n_2$ attributes of the $i$-th category.

Then we obtain the text descriptions 
of each category by asking additional questions. Note that these text descriptions of attributes  may contain noise, we call them coarse attribute descriptions. This process is formulated by
\begin{equation} \label{eq:ask}
    \mathbb{D}_i^{c} = \Phi(\Pi_2(\mathbb{A}_c \cup \mathbb{A}_{s,i}, Y_i)),
\end{equation}
where $\Pi_2(\cdot)$ denotes the questions about describing the attributes of objects, like \textit{please help me generate 100 different sentences about [category] from the angle of the [attribute]}, and  $\mathbb{D}_i^{c}$ denotes the coarse attribution descriptions of the $i$-the category. Let  $\mathbb{D}^{c} =\{\mathbb{D}_1^{c},\cdots,\mathbb{D}_N^{c}\}$  denote the coarse attribute description sets. 
%Note that since the attributes for each object may contain noise, we call the corresponding descriptions $\mathbb{D}^{c} =\{\mathbb{D}_1^{c},\cdots,\mathbb{D}_N^{c}\}$ as 

% To get diverse aspects of objects, we inquire LLMs questions for each category, like ``give me $M_1$ attribute words that are relevant to the [category$_1$, $\cdots$, category$_N$]", $M_1$ attributes can be generated for all category, and then we ask question like `` besides 'attribute$_1$ $\cdots$, attribute$_{M_1}$', generate another $M_2$ attributes for [category]" for each category, resulting another $M_2$ attributes for each category. 
% The final attribute list for each category has $M=M_1+M_2$ items. And then for each category and for each attribute word, we ask the question ``generate 10 sentences to describe the [attribute] of [category]", the resulting sentences as well as the category name itself served as coarse attribute knowledge.

\noindent \textbf{Fine-grained Attribute Description.}
We design several questions to remove the noisy attributes that are irrelevant to the specific category, resulting in a fine-grained attribute set for each category. We then inquire ChatGLM to acquire the fine-grained attribute descriptions $\mathbb{D}^{f} = \{\mathbb{D}^{f}_1,\cdots,\mathbb{D}^{f}_N\}$ by asking questions $\Pi_2(\cdot)$ of Eq.(\ref{eq:ask}). This process is formulated by
\begin{equation} \label{eq:fine_grained}
    \mathbb{D}^{f}_i =\Phi(\Pi_2(\Phi(\Pi_3(\mathbb{A}_c \cup \mathbb{A}_{s,i}, Y_i))),Y_i),
\end{equation}
where $\Pi_3(\cdot)$ denotes the questions about how to remove irrelevant attributes, like \textit{[attribute list], please delete the above attribute words given that are not very relevant to [category]. Finally, 70 attribute words remain}.
% Then, based on the fine-grained attribute set $\mathring{\mathbb{A}}=\{\mathring{\mathbb{A}}_i,\cdots,\mathring{\mathbb{A}}_N\}$, asking the questions of $\Pi_2(\cdot)$ of Eq.(\ref{eq:ask}) we can get fine-grained attributes description $\mathbb{D}^{f}$.
% As the attribute list for each category may have noisy items, before we generate the text to describe the attributes, we first remove the items that are not relevant to the categories by asking the question ``is the [attribute] relevant to the [category]?" repeatedly for all attributes of each category. 
% And then the more clean attribute lists lead to fine-grained attributes knowledge.

\noindent \textbf{Relationship Description.}
In multi-label image recognition,  the co-occurrence relationships between different categories contribute significantly to the performance~\cite{chen2019multi,wang2020multi}.
To simulate this scenario, we first split all categories into multiple scene-related subgroups by ChatGLM:
\begin{equation}
    \mathbb{G} = \{\mathbb{G}_i\}_{i=1}^{n_3} = \Phi(\Pi_4(\mathbb{Y})),
\end{equation}
where $\Pi_4(\cdot)$ denotes questions about how to divide categories into  subgroups based on their relationships, like \textit{[category list], categorize the above words according to possible common occurrences in a scene}, and $n_3$ is the number of subgroups.
For each subgroup, two categories are selected to formulate scene-related questions $\Pi_5(\cdot)$, like \textit{generate 100 different descriptive sentences for a scene containing [category1] and [category2]},  and fed into ChatGLM to obtain relationship descriptions: %$\mathbb{D}^{r}=\{\mathbb{D}_1^r,\cdots,\mathbb{D}_N^r\}$:
\begin{equation} \label{eq:relationships}
    \mathbb{D}_{i}^{r} = \Phi(\Pi_5(\mathbb{G}_i)), 
\end{equation}
where $\mathbb{D}_{i}^{r} $ denotes the fine-grained attribute descriptions of the $i$-th categroy. Let $\mathbb{D}^{r}=\{\mathbb{D}_1^r,\cdots,\mathbb{D}_N^r\}$ denote the fine-grained attribute description sets.

\begin{figure}
    \centering
    \includegraphics[width=\linewidth]{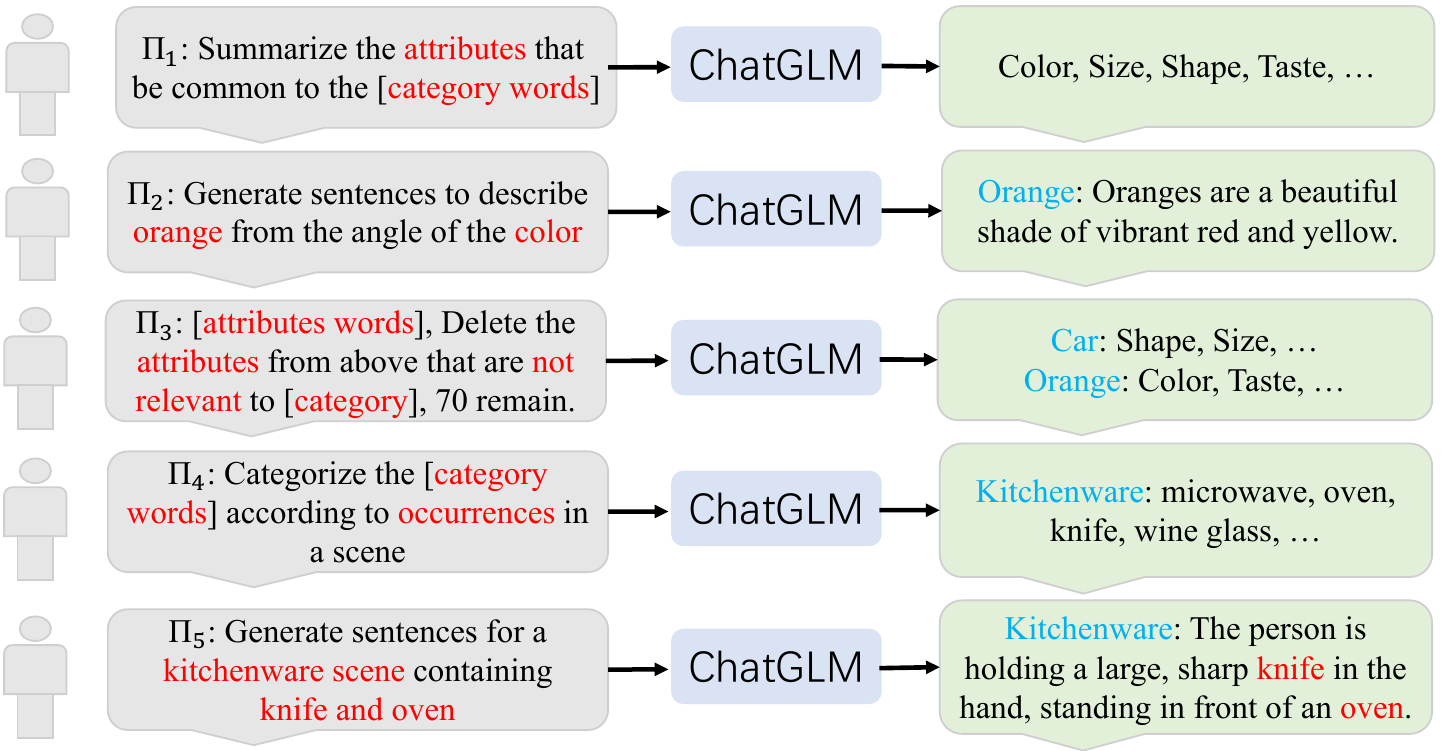}
    \caption{An example of the designed questions and their corresponding answers from ChatGLM. More detailed examples can be found in the supplementary materials.}
    \label{fig:questions}
\end{figure}

Figure~\ref{fig:questions} illustrates an example of the designed questions and their corresponding answers from ChatGLM.

\subsection{Hierarchical Prompt Learning }
Based on the previous generated text descriptions $\mathbb{D} = \mathbb{D}^k, k\in\{c,f,r\}$, 
we propose hierarchical prompt learning to adapt CLIP to multi-label recognition, where hierarchical prompts are designed to model relationships between categories, and both global and local prompt learning are introduced to grasp the discriminability of features.
%similar to TaI-DPT~\cite{guo2023texts}, we conduct a global learning and local learning that learn both global prompts and local prompts to grasp the discriminability of global features as well as fine-grained features, both prompts are hierarchical prompts that consider the relationships between categories.

\noindent \textbf{Hierarchical Prompts.}
A learnable prompt usually consists of several learnable tokens and a placeholder to put the category label, denoted as  $\mathbf{p}_i=[\mathbf{t}_1^i,\mathbf{t}_2^i,\cdots,\mathbf{t}_M^i, Y_i]$, where $M$ is the number of learnable tokens, and $Y_i$ is the $i$-th category label.
For different categories, there are two types of prompts: shared prompts, where tokens are shared across different categories, and category-specific prompts, where tokens are distinct for each category.
%For all categories, we design two types of prompts: (1) shared tokens version $\mathbf{P}^s$,
% \begin{equation}
%     \mathbf{P}^s = \begin{bmatrix}
%   \mathbf{t}_1, & \mathbf{t}_2, & \mathbf{t}_3, & \cdots & \mathbf{t}_\Omega, & Y_1\\
%   % \mathbf{t}_1, & \mathbf{t}_2, & \mathbf{t}_3, & \cdots & \mathbf{t}_\Omega, & Y_2 \\
%   & \ddots &  & \ddots &  & \\
%   \mathbf{t}_1, & \mathbf{t}_2, & \mathbf{t}_3, & \cdots & \mathbf{t}_\Omega, & Y_N
% \end{bmatrix},
% \end{equation}
% where learnable tokens are shared over all categories, the only difference is the category name; and (2) category-specific tokens version  $\mathbf{P}^s$,
% \begin{equation}
%     \mathbf{P}^c = \begin{bmatrix}
%   \mathbf{t}_1^1, & \mathbf{t}_2^1, & \mathbf{t}_3^1, & \cdots & \mathbf{t}_\Omega^1, & Y_1\\
%   % \mathbf{t}_1^2, & \mathbf{t}_2^2, & \mathbf{t}_3^2, & \cdots & \mathbf{t}_\Omega^2, & Y_2 \\
%   &  \ddots &  & \ddots &  & \\
%   \mathbf{t}_1^N, & \mathbf{t}_2^N, & \mathbf{t}_3^N, & \cdots & \mathbf{t}_\Omega^N, & Y_N
% \end{bmatrix},
% \end{equation}
% where learnable tokens for different categories are specific. 
Both types of prompts have been demonstrated effective in recent works~\cite{zhou2022learning,zhu2023prompt}, but they neglect the relationships of different categories, leading to sub-optimal performance.

To capture the relationships between different categories, we propose hierarchical prompts $\mathbf{P}^h$, a mixed version of both shared tokens and category-specific tokens, and additional partial-shared tokens only between categories that most likely co-occur in a scene or have similar attributes, denote as
\begin{equation*} \label{eq:hierarchical}
  \mathbf{P}^h = \begin{bmatrix}
   \mathbf{t}_{s}, &  \mathbf{t}_{p,1}, & \mathbf{t}_{p,2}, & \mathbf{t}_{p,3}, & \cdots & \mathbf{t}_{c,1}, & Y_1 \\
   \mathbf{t}_{s}, &  \mathbf{t}_{c,a}, & \mathbf{t}_{p,2}, & \mathbf{t}_{p,4}, & \cdots & \mathbf{t}_{c,2}, & Y_2 \\
   \mathbf{t}_{s}, &  \mathbf{t}_{p,1}, & \mathbf{t}_{p,2}, & \mathbf{t}_{p,3}, & \cdots & \mathbf{t}_{c,3}, & Y_3 \\
       &  \ddots &  &     &  \ddots&   \\
   \mathbf{t}_{s}, &  \mathbf{t}_{p,1}, & \mathbf{t}_{c,b}, & \mathbf{t}_{p,4}, & \cdots & \mathbf{t}_{c,N}, & Y_N 
\end{bmatrix},
\end{equation*}
where all tokens are divided into three types: (1) shared tokens, denoted as $\mathbf{t}_{s}$, that are shared over all categories; (2) partial-shared tokens, denoted as $\mathbf{t}_{p,\varsigma}, \varsigma \in \{1,2,3,4\}$, that are tokens shared only between some categories; (3) category-specific tokens, denoted as $\mathbf{t}_{c,\varepsilon},\varepsilon \in \{a,b,1,\cdots,N\}$, that are distinct for different categories. Note that for the columns of partial-shared, \textit{i.e.}, the columns 2,3,4 of $\mathbf{P}^h$, %in Eq.(\ref{eq:hierarchical}), 
there are also some category-specific tokens ($\mathbf{t}_{c,a}$ and $ \mathbf{t}_{c,b}$) due to no relationships in that circumstance. For example, in the kitchenware scene, the knife and oven have a shared token, but the corresponding tokens of the sofa and book may be specific.
% $\mathbf{t}_1$ is the shared tokens, $\mathbf{t}_j, j\in \{2,3,4,c\}$ denotes partial-shared tokens and $\mathbf{t}_a, \mathbf{t}_b, \mathbf{t}_\Omega^k, k\in\{1,\cdots,N\}$ are category-specific tokens.
% where we split the tokens of prompts into three types: (1) shared tokens ($\mathbf{t}_1$), that are shared over all categories;  (2) partial-shared tokens ($\mathbf{t}_j, j\in \{2,3,4,c\}$), that are only shared between some of the categories ;(2) category-specific tokens ($\mathbf{t}_a, \mathbf{t}_b, \mathbf{t}_\Omega^k, k\in\{1,\cdots,N\}$ ).

% More specifically, the partially shared tokens are divided into two levels, a fine-grained level that groups categories by asking the LLMs ``xxx", and then a coarse level that groups the find-grained level again by asking the LLMs ``xxx". An illustration of graph prompts can be found in Figure~\ref{fig:graph}, more details are provided in supplementary materials.

% To learn the hierarchical prompts, similar to the previous methods~\cite{zhou2022conditional}, we maximize the probability of classifying each image $x$ into its ground-truth class $y$:
% \begin{equation}
%     p(y=y_i|\mathbf{x}) = \frac{\text{exp}(\left <\Psi_i(\mathbf{x}),\Psi_t(\mathbf{p_i^g}\right >)/\tau)}{ {\textstyle \sum_{j=1}^{N}}\text{exp}(\left <\Psi_i(\mathbf{x}),\Psi_t(\mathbf{p_j^g}\right >)/\tau},
% \end{equation}
% where $p_i^g$ is the prompts of categories$_i$; $\left <\cdot,\cdot\right >$ denotes the cosine similarity and $\tau$ is the temperate hyper-parameter.

\noindent \textbf{Global Learning.}
Global learning aims to learn global hierarchical prompts to grasp the discriminative ability of global features. Let $\mathbf{P}_{g,i}^h$ be the global hierarchical prompts of the $i$-th category. 
%denote as $\mathbf{P}_g^h = [\mathbf{P}_{g,1}^h,\mathbf{P}_{g,2}^h, \cdots,\mathbf{P}_{g,N}^h]$, 
We initialize $\mathbf{P}_{g,i}^h$ randomly  and then feed it to the text encoder $\Psi_t$ of CLIP to generate the global category embedding $\mathbf{G}_i$:
\begin{equation} \label{eq:global_class_embedding}
    \mathbf{G}_i = \Psi_t(\mathbf{P}_{g,i}^h) \in \mathbb{R}^{d}, \quad i \in \{1,2,\cdots,N\},
\end{equation}
where $d=512$ denotes the embedding dimension.
Meanwhile, for each text description $\mathbf{r} \in \mathbb{D}$, we extract its global feature $\mathbf{T}^G$:% that are the output at location of $\left <EOS\right> $ token: 
\begin{equation}
    \mathbf{T}^G = \Psi_t(\mathbf{r}) \in \mathbb{R}^{d}, \quad \mathbf{r} \in \mathbb{D}.
\end{equation}
The cosine similarity between the global feature of text description  and the global category embedding, denoted as $S_i^G$, is calculated by %$ \mathbf{S}^G=[S_1^G,\cdots,S_N^G] \in \mathbb{R}^{N}$ :
\begin{equation}\label{eq:global}
    S_i^G = \left <\mathbf{T}^G, \mathbf{G}_i \right >, i\in\{1,\cdots,N\}.
    % \mathbf{S}^G = [S_1^G,\cdots, S_N^G], \quad S_i^G = \left <\mathbf{T}^G, \mathbf{G}_i \right >, i\in\{1,\cdots,N\}.
\end{equation}

\noindent \textbf{Local Learning.}
Local learning aims to learn local hierarchical prompts to grasp the discriminative ability of  fine-grained features.
Let $\mathbf{P}_{l,i}^h$ be the local hierarchical prompts of the $i$-th category, whose structure is identical to global prompts but with different parameters. We initialize $\mathbf{P}_{l,i}^h$ randomly and feed it to the encoder $\Psi_t$ of CLIP to generate the local category embedding $\mathbf{L}_i$:
%We initialize local hierarchical prompts randomly, denote as $\mathbf{P}_l^h = [\mathbf{P}_{l,1}^h,\mathbf{P}_{l,2}^h, \cdots,\mathbf{P}_{l,N}^h]$ for all categories, whose structure are identical to global prompts but with different parameters,  and then feed them to the text encoder $\Psi_t$ of CLIP to generate local class embedding:
\begin{equation} \label{eq:local_class_embedding}
    \mathbf{L}_i = \Psi_t(\mathbf{P}_{l,i}^h) \in \mathbb{R}^{d}, \quad i \in \{1,2,\cdots,N\}.
\end{equation}
For the text description $\mathbf{r} \in \mathbb{D}$, we extract its local features $\mathbf{T}_i^L$ by a modified text encoder $\widetilde{\Psi}_t$:
\begin{equation}
    \mathbf{T}^L = \widetilde{\Psi}_t(\mathbf{r}) \in \mathbb{R}^{N_r \times d}, \quad \mathbf{r} \in \mathbb{D},
\end{equation}
where $N_r$ is the number of tokens in $\mathbf{r}$, $\widetilde{\Psi}_t$ denotes that we preserve the sequential token features of the entire sentence instead of only the $\left <EOS\right> $ token features (global features).
%zA category-aware local similarity $S^L = [S_1^L, \cdots,S_N^L] \in \mathbb{R}^N$ between sequential text features and local class embedding is calculated in a weighted manner: 
The category-aware  similarity between the sequential local features of text description and the local category embedding is calculated in a weighted manner: 
\begin{equation}\label{eq:local}
    S_i^L =  {\textstyle \sum_{j=1}^{N_r}} \frac{\text{exp}(\mathbf{s}_{i,j})}{{\textstyle\sum_{j=1}^{N_r}}\text{exp}(\mathbf{s}_{i,j})} \cdot \mathbf{s}_{i,j}, \quad i\in\{1,\cdots,N\},
\end{equation}
where $\mathbf{s}_{i,j} = \left <\mathbf{L}_i, \mathbf{T}_j^L \right >$ is the similarity between the $i$-th local class embedding and $j$-th token (column) of local features. 

\section{Training Objectives}
For the global similarity $\mathbf{S}^G = [S_1^G,\cdots, S_N^G]$ and local similarity $\mathbf{S}^L = [S_1^L,\cdots, S_N^L]$ of each text description, we adopt two loss functions to optimize the corresponding learnable prompts.

\noindent \textbf{Ranking Loss.}
We utilize the ranking loss to assess the disparity between classification scores and ground-truth labels. Specifically, the  ranking loss for the global and local learning is calculated separately: 
\begin{equation}
    \begin{aligned}
        &\mathcal{L}_{rank} = \mathcal{L}_{global} + \mathcal{L}_{local}, \\
        & \mathcal{L}_{global} = {\textstyle \sum_{i \in \mathbb{Y}^{+},j \in \mathbb{Y}^{-}}}\text{max}(0, m-S_{i}^{G}+S_{j}^{G})&, \\
        & \mathcal{L}_{local} = {\textstyle \sum_{i \in \mathbb{Y}^{+},j \in \mathbb{Y}^{-}}}\text{max}(0, m-S_{i}^{L}+S_{j}^{L})&, \\
 \end{aligned}
\end{equation}
where $m=1$ is the margin controlling how much higher the similarity score with the positive classes $\mathbb{Y}^{+}$ is than with the negative classes $\mathbb{Y}^{-}$. 
%The overall ranking loss can be formulated as
%\begin{equation}
   
%\end{equation}

\noindent \textbf{Order Loss.} 
Due to the potential noise introduced by text generated from ChatGLM, which could mislead the optimization process, we introduce an anchor to the learned prompts. Specifically, we anchor the learned prompts using hand-craft prompts, such as \textit{a photo of [category]}. The rationale behind this is that human-defined prompts have demonstrated mediocre zero-shot performance,  thus the resulting order of all given categories can be considered reasonable to some extent. We posit that maintaining this order in the learned prompts helps mitigate the impact of noisy inputs, thereby enhancing the overall effectiveness of the model.

\begin{table*}
\centering
\caption{Results of different text descriptions on MS-COCO, VOC2007 and NUS-WIDE. $\dagger$ means that results are from TaI-DPT~\cite{guo2023texts}. Note that the hand-craft prompts used in this paper are designed specifically for each category.}
\scalebox{1}{\begin{tabular}{l|c|cc|cc|cc}
\hline \hline
\multicolumn{1}{c|}{\multirow{2}{*}{Knowledge}} & \multirow{2}{*}{Training}& \multicolumn{2}{|c|}{MS-COCO} & \multicolumn{2}{|c}{VOC2007} & \multicolumn{2}{|c}{NUS-WIDE} \\
     && \multicolumn{1}{c}{F1} & \multicolumn{1}{c|}{mAP} & \multicolumn{1}{c}{F1} & \multicolumn{1}{c|}{mAP} & \multicolumn{1}{c}{F1} & \multicolumn{1}{c}{mAP} \\ \hline
\multicolumn{1}{l|}{Hand-craft prompts$^{\dagger}$ (zero-shot)} & $\times$ & -  & 49.7 & - & 77.3 & - & 37.4 \\ 
\multicolumn{1}{l|}{Hand-craft prompts  (zero-shot)} & $\times$ & 47.7  & 62.1 & 58.5 & 86.6 & 30.4 & 43.3 \\ \hline
\multicolumn{1}{l|}{Image Captions} & $\checkmark$& 48.8  & \textbf{67.5} & 56.1 & 86.8 & \textbf{43.4} & 44.1\\ \hline
\multicolumn{1}{l|}{Coarse Attribute} & $\checkmark$ & 51.9  & 64.2 & 57.1 & 88.0 & 36.6 & 46.0\\
\multicolumn{1}{l|}{Fine-grained  Attribute } & $\checkmark$ & 52.1  & 64.4 & 58.1 & 88.2 & 36.8 & 46.1\\ %& 36.11 & 45.87
\multicolumn{1}{l|}{Ours} & $ \checkmark$ & \textbf{57.3}  & 66.8 & \textbf{60.0} & \textbf{88.7} & 40.0 & \textbf{47.0}\\ \hline
\hline
\end{tabular}}
\label{tab:abl_dataset}
\end{table*}	

\begin{table}[t]
\centering
\caption{Results of different prompts on  MS-COCO.}
\scalebox{1}{\begin{tabular}{l|cc}
\hline \hline
\multicolumn{1}{c|}{Prompts} & \multicolumn{1}{|c}{F1} & \multicolumn{1}{c}{mAP} \\ \hline
\multicolumn{1}{l|}{Hand-craft}   & 47.7 & 62.1    \\
\multicolumn{1}{l|}{Category-specific}   & 52.7 & 64.8    \\
\multicolumn{1}{l|}{Shared}   & 54.7 & 66.3   \\
\multicolumn{1}{l|}{Hierarchical (Ours)}   & \textbf{57.3} & \textbf{66.8}   \\
\hline  \hline
\end{tabular}}
\label{tab:abl_prompts}
\end{table}

For learnable global and local prompts and hand-craft prompts, denoted as $G, L, H$, we extract their category embedding using Eq.(\ref{eq:global_class_embedding}), denoted by $\mathbf{G}_k, k \in \{G,L,H\}$, followed by similarity calculation between different categories: $\mathbf{D}^{k} = \mathbf{G}_k \times \mathbf{G}_k^{\top}, k \in \{G,L,H\}$.
% resulting in an order representation, which is the distribution of similarities.
% For an input text description, we replace the learnable hierarchical prompts with the hand-craft prompts, such as "a photo of [category]", and calculated the global similarities ${S}^{G'}_i$ and local similarities ${S}^{L'}_i$ by Eq.(\ref{eq:global}) and Eq.(\ref{eq:local}). 
The order loss is then calculated by the Kullback-Leibler (KL) divergence :
% \begin{equation} \label{eq:loss_order}
%     \mathcal{L}_{order} = \text{KL}(\mathbf{S}^{L},\mathbf{S}^{L'}) +  \text{KL}(\mathbf{S}^{G},\mathbf{S}^{G'}), 
% \end{equation}
% where $\mathbf{S}^{G'} = [S_1^{G'},\cdots, S_N^{G'}]$ and $\mathbf{S}^{L'} = [S_1^{L'},\cdots, S_N^{L'}]$.
\begin{equation} \label{eq:loss_order}
    \mathcal{L}_{order} = \text{KL}(\mathbf{D}^{G},\mathbf{D}^{H}) +  \text{KL}(\mathbf{D}^{L},\mathbf{D}^{H}), 
\end{equation}

Finally, the overall loss is given by 
\begin{equation} \label{eq:loss}
    \mathcal{L} = \mathcal{L}_{rank} + \lambda_1 \cdot \mathcal{L}_{order}
\end{equation}

\subsection{Inference}
Thanks to large-scale image-text contrastive pre-training of CLIP,  text features have been well-aligned to the image features of the same semantic meanings. 
As a result, our prompts learned from  text descriptions can be applied to images during inference.  

Specifically, with the learned hierarchical prompts, we extract the global and local class embedding using Eq.(\ref{eq:global_class_embedding}) and Eq.(\ref{eq:local_class_embedding}) repeatly, denoted as $\mathbf{T}^{G}=[\mathbf{T}_1^{G},\cdots,\mathbf{T}_N^{G}]$ and $\mathbf{T}^{L}=[\mathbf{T}_1^{L},\cdots,\mathbf{T}_N^{L}]$. 
For an input image $\mathbf{x}$, we extract the global and local image features by 
\begin{equation}
    \mathbf{I}^{G} = \Psi_i(\mathbf{x}) \in \mathbb{R}^{d}, \quad \quad  \mathbf{I}^{L} = \widetilde{\Psi}_i(\mathbf{x}) \in \mathbb{R}^{N_I \times d},
\end{equation}
where $\mathbf{I}^{G}$ is the global image feature, $\mathbf{I}^{L}$ is the local image feature, $\widetilde{\Psi}_i$ is the modified image encoder of CLIP that keeps the dense image
features as output, and $N_I$ is the length of flatted dense image features.
Finally, we calculate the similarities between category embeddings and image features by 
\begin{equation} \label{eq:inference}
   \mathbf{S}= \lambda_2 \cdot \mathbf{S}^{G} + (1-\lambda_2) \cdot \mathbf{S}^{L},
\end{equation}
where $\lambda_2$ is a parameter to weigh how much contribution of the global prompts and local prompts. $\mathbf{S}^{G} =[ \left< \mathbf{I}^{G},\mathbf{T}_1^{G} \right>,\cdots,\left< \mathbf{I}^{G},\mathbf{T}_N^{G} \right>]$ is the global similarity. $\mathbf{S}^{L} = [{S}_1^{L},\cdots,{S}_N^{L}]$ is the local similarity, where
\begin{equation*}
    {S}_i^{L}={\textstyle \sum_{j=1}^{N_I}} \frac{\text{exp}(\mathbf{s}_{i,j}/\tau)}{{\textstyle\sum_{j=1}^{N_I}}\text{exp}(\mathbf{s}_{i,j}/\tau)} \cdot \mathbf{s}_{i,j},
\end{equation*} 
and $\mathbf{s}_{i,j} = \left<\mathbf{T}_i^{L}, \mathbf{I}_j^{L}\right>$ is the cosine similarity between the $i$-th local category embedding and the $j$-th token (column) of local image features.

% For multi-label image recognition, similar to TaI-DPT~\cite{guo2023texts}, we learn two types of prompts, a global prompts for utilizing the whole image information, and a local prompts to force the model look at local regions.
% what's more, to utilize the dependency between all categories, 

\section{Experiments}
\label{sec:exper}
\subsection{Datasets and Evaluation Metrics}
\noindent \textbf{Datasets}.
We conduct experiments on the MS-COCO~\cite{lin2014microsoft}, VOC2007~\cite{everingham2010pascal} and NUS-WIDE~\cite{chua2009nus} datasets for evaluation. For all three datasets, no training data is used and the testing is performed on the testing or validation sets. % (annotation of the testing set is unavailable), and . 
MS-COCO is a widely used multi-label dataset for image recognition, which contains 80 categories with 82,081 training images and 40,504 validation images. VOC2007 contains 20 object categories with a total of 5,011 images for training and validation and 4,952 images for testing. NUS-WIDE  contains 81 categories with 161,789 images for training and 107,859 images for testing.

\noindent \textbf{Metrics}. 
We use the conventional  evaluation metrics, including the mean of class-average precision (mAP) and the overall F1 score at Top-3 predictions.

\subsection{Implementation Details}
% The pre-trained large language model (LLM) used in the paper is v1.0 of ChatGLM~\cite{du2022glm}.
% We adopt ResNet-50~\cite{radford2021learning} as the visual encoder of CLIP for input resolution 224 $\times$ 224, and the pre-trained CLIP is kept frozen during optimization. 
% We learn two independent hierarchical prompts with 32 context tokens, while a longer sequence brings trivial improvements. 
% The learning rates for all datasets are empirically initialized with 0.002, decayed by 0.1 when epochs at 2 and 5, and total training epochs are set to 10.
% The learning rates for MS-COCO, VOC2007, and NUS-WIDE are empirically initialized with 0.002, 0.002, and xx, decayed by the 0.1 when epoch 2 and 5. 
We employ ResNet-50 as the visual encoder of CLIP with an input resolution of 224 $\times$ 224. 
In prompt learning, the number of context tokens  in hierarchical prompts is set to 32. 
The learning rate is set to 0.002 with a decay of 0.1 at the epochs of 2 and 5, and the total training epoch is set to 10.
The prarmeters $\lambda_1$ in Eq.(\ref{eq:loss}) and $\lambda_2$ in Eq.(\ref{eq:inference}) are set to 0.2 and 0.65, respectively. 

\subsection{Ablation Studies}
%In this section, we perform in-depth analysis to evaluate each component of our method.

% \subsubsection{Ablation study on textual generation}
\noindent \textbf{Effectiveness of different text descriptions.} 
To evaluate the acquired text descriptions from ChatGLM by our method, we employ  different inquiry strategies  to obtain  text descriptions for comparison, including (1) ``Hand-craft prompt": the inference is directly performed using hand-craft prompts without prompt tuning; (2) ``Image Captions": the human-written image captions from existing datasets are used for prompt tuning; (3) ``Coarse Attribute": the text descriptions of object attributes with noise are used for prompt tuning, generated by Eq.(\ref{eq:coarse}); (4) ``Fine-grained Attribute": the text descriptions of filtered object attributes are used for prompt tuning, % denotes that the training data of prompt tuning is descriptions of , 
generated by Eq.(\ref{eq:fine_grained}). %(5)``Relationship Descriptions" denotes that the training data of prompt tuning is descriptions of object relationships. 

Table~\ref{tab:abl_dataset} shows the results of different text descriptions on the MS-COCO, VOC2007 and NUS-WIDE datasets.
% It is obvious that prompt-tuning based on the generated dataset can significantly improve performance, about 10\% gains from 57.87\% to 67.54 \%,  demonstrating the knowledge from LLMs helps a lot to boost the VLMs's performance on multi-label image recognition by prompt-tuning learning.
% Specifically, inquiring about the characteristics (attributes) of categories from LLMs, prompts learned based on both the ``Coarse" dataset and its clean version ``Fine-grained" dataset,  improve the performance of the human-define prompt ``None" by more than 6\% and prompts learned based on ``Scene" dataset further gains 2.44\% considering the relationships of different categories.
% Moreover, 1.2\% gains of mAP can be further achieved using the in-domain text knowledge, \textit{i.e.}, the image caption of the training dataset,  which means the generated dataset contains noisy information. 
We have the following observations: (1) Our method achieves  substantial improvements over ``Hand-crafted prompts", with F1 score gains of 7.64\%, 1.55\%, and 9.6\% on the three datasets, respectively. This highlights the remarkable contribution of knowledge extracted from ChatGLM in enhancing the zero-shot performance of CLIP on multi-label image recognition;
(2) Our method outperforms ``Fine-grained Attribute" descriptions by approximately 5.2\%, 1.9\% and 3.2\% in F1 score on the three datasets, respectively. This superiority emphasizes that considering the relationships between objects captures more discriminative information to enhance multi-label recognition;
(3) Our method performs better than ``Image Captions" on most metrics, suggesting that ChatGLM provides more comprehensive knowledge than human-written image captions; 
(4) The performance of ``Coarse Attribute" is lower than that of ``Fine-grained Attribute", confirming that the presence of noise within the coarse attributes of objects degrades the performance.

\noindent \textbf{Effectiveness of different prompts.}
% Different prompt methods are evaluated in Table~\ref{tab:abl_prompts}. ``Hand-craft" means human-defined prompts like ``an image of [category]", and we customize prompts for each category; ``Shared" denotes learnable tokens of prompts are shared across all categories; ``category-specific" denotes tokens of prompts are specific for each category; and ``Hierarchical" denotes our relationship-aware partial-shared prompts, that tokens are shared only between some categories. 
To evaluate the proposed hierarchical prompts, we employ different types of prompts for comparison, including (1) ``Hand-craft": the human-defined prompts are customized for each category like \textit{an image of [category]}; (2) ``Category-specific":  the learnable tokens of prompts are specific for each category; (3) ``Shared": the learnable tokens of prompts are shared across all categories.
% Table~\ref{tab:abl_prompts} presents their results on NS-COCO dataset. The term ``Hand-craft" refers to human-defined prompts, such as \textit{an image of [category]}, tailored individually to each category. ``Shared" signifies learnable prompt tokens shared across all categories, ``Category-specific" indicates tokens specific to each category, while ``Hierarchical" signifies our relationship-aware hierarchical prompts.

Table~\ref{tab:abl_prompts} shows the results of different types of prompts on MS-COCO.
%From Table~\ref{tab:abl_prompts}, 
Our hierarchical prompts outperform all other methods, demonstrating the effectiveness of incorporating inter-category relationships into prompts. Moreover, compared to the performance of hand-crafted prompts, that of learnable prompts (i.e., category-specific, shared and hierarchical prompts) is much higher. Interestingly, ``Shared prompts" outperforms ``Category-specific",  indicating the better generalization ability of shared prompts.

%Surpassing all other methods, our hierarchical prompts method showcases its effectiveness in leveraging inter-category relationships for designing a more effective prompt structure. 
%Indeed, the performance of the hand-crafted prompts is much lower than that of learnable prompts by 2.68\%, 4.2\% and 4.66\% in terms of mAP on the MS-COCO dataset, demonstrating the superiority of learnable prompts.
%Interestingly, the performance of the ``Category-specific" prompts is inferior to the ``Shared prompts," indicating the better generalization ability of shared prompts.

% Though we specifically designed hand-craft prompts for each object category, the performance is much lower than that of learnable prompts by 4.66\% in terms of mAP in the MS-COCO dataset.
% And the hierarchical prompts method outperforms all others, demonstrating the effectiveness of considering inter-category relationships when designing the proper prompt structure. 
% And the performance of ``Class-specific" prompts is inferior to that of ``Shared prompts", indicating that the shared prompt method generalizes better than the class-specific prompt method.  

\begin{table}[t]
\centering
\caption{Results of different components on  MS-COCO.}
\label{tab:abl_component}
\begin{tabular}{cc|cc}
\hline \hline
\multicolumn{1}{c}{Local learning} & \multicolumn{1}{c|}{Order loss} & \multicolumn{1}{c}{F1}& \multicolumn{1}{c}{mAP}\\ \hline 
 $\times$ &  $\times$          &        52.4          & 62.2 \\ 
 \checkmark &  $\times$         &      56.3     & 66.0 \\
 $\times$ & \checkmark           &    53.2     & 63.2 \\
 \checkmark & \checkmark & \textbf{57.3}  &  \textbf{66.8}\\	
\hline \hline
\end{tabular}
\end{table}

\noindent \textbf{Effectiveness of order loss.}
%We proposed an order loss (Eq.(\ref{eq:loss_order})) to align the outputs of the learning prompts with those hand-craft prompts, aiming to mitigate the potential impact of noisy information from the generated textual knowledge. 
To evaluate the effectiveness of the order loss in Eq.(\ref{eq:loss_order}), we remove it for comparison. The results on MS-COCO are shown in  Table~\ref{tab:abl_component}, verifying the advantage of the order loss in  mitigating the  noisy of  text descriptions acquired from ChatGLM. 

\noindent \textbf{Effectiveness of local learning.}
%We introduce local learning to force the CLIP to focus on sub-regions of inputs. 
%We evaluate its effectiveness in Table~\ref{tab:abl_component}.
%When comparing line 1 with line 2, and line 3 with line 4, the local learning improves more than 3\% in mAP on the MS-COCO dataset, underscoring the significant impact of concentrating on the sub-region of image in multi-label image recognition.
To evaluate the effectiveness of the local learning of hierarchical prompts, we remove it for comparison. The results on MS-COCO are shown in  Table~\ref{tab:abl_component}, highlighting the significant impact of focusing on image sub-regions in multi-label  recognition.

% In the training objective, we incorporate an order loss (Eq.(\ref{eq:xxxkl})) to ensure that the  outcomes of the learning prompts align closely with those of hand-craft prompts.
% This is because there might be noisy information of generated textual knowledge, which may mislead the optimizations.
% The corresponding results are presented in Table~\ref{tab:abl_component}. When comparing line 1 with line 3, and line 2 with line 4, adding the order loss to the framework yields gains of 0.94\% and 0.77\%, respectively, highlighting its positive impact.

\begin{table}[t]
\centering
\caption{Results of different numbers of different token types in hierarchical prompts on MS-COCO. S: shared token; P-S \#1: partial-shared token over more categories (within coarse subgroups); P-S \#2: partial-shared token over fewer categories (within more fined subgroups); C-S: category-specific token.}
\label{tab:abl_mix_tokens}
\begin{tabular}{cccc|cc}
\hline
\hline
\multicolumn{1}{c}{S} & \multicolumn{1}{c}{P-S \#1} & \multicolumn{1}{c} {P-S \#2}&\multicolumn{1}{c}{C-S}& \multicolumn{1}{|c}{F1} & \multicolumn{1}{c}{mAP} \\ \hline 
8 & 8 & 8 & 8 & 56.4 & 66.6\\
12 & 8 & 6 & 6 & 57.2 & 66.6\\
16 & 8 & 4 & 4 & \textbf{57.3} & \textbf{66.8}\\
20 & 4 & 4 & 4 & 57.1 & 66.7\\
20 & 6 & 4 & 2 & 56.0 & 66.5\\
\hline
\hline
\end{tabular}
\end{table}	

\begin{table*}
\centering
\caption{Comparison results (mAP) with the state-of-the-art methods on MS-COCO, VOC2007 and NUS-WIDE.}
\begin{tabular}{l|c|c|c|ccc}
\hline \hline
Method                                                          & Venue        & Training source        & Annotation                         & MS-COCO & VOC2007 & NUS-WIDE \\ \hline
SRN~\cite{zhu2017learning}                & CVPR 2017    & \multirow{3}{*}{Image} & \multirow{3}{*}{Fully labeled}     & 77.1    & -       & 62.0     \\
ML-GCN~\cite{chen2019multi}               & CVPR 2019    &                        &                                    & 83.0    & 94.0    & -        \\
ASL~\cite{ridnik2021asymmetric}           & ICCV 2021    &                        &                                    & 86.6    & 94.6    & 65.2     \\ \hline
SARB~\cite{pu2022semantic}               & AAAI 2022    & \multirow{3}{*}{Image} & \multirow{3}{*}{Partially labeled} & 71.2    & 83.5    & -        \\
SST~\cite{chen2022structured}             & AAAI 2022    &                        &                                    & 68.1    & 81.5    & -        \\
DualCoOp~\cite{sun2022dualcoop}           & NeurIPS 2022 &                        &                                    & 78.7    & 90.3    & -        \\ \hline
LL-R~\cite{kim2022large}                 & CVPR 2022    & \multirow{2}{*}{Image} & \multirow{2}{*}{One labeled}       & 72.6    & 90.6    & 47.4     \\
G$^2$NetPL~\cite{abdelfattah2022g2netpl} & BMVC 2022    &                        &                                    & 72.5    & 89.9    & 48.5     \\ \hline
LSAN~\cite{szegedy2016rethinking}         & CVPR 2016    & \multirow{4}{*}{Image} & \multirow{4}{*}{Unlabeled}         & 65.5    & 87.9    & 41.3     \\
WAN~\cite{mac2019presence}                & ICCV 2019    &                        &                                    & 63.9    & 86.2    & 40.1     \\
Curriculum~\cite{durand2019learning}      & CVPR 2019    &                        &                                    & 63.2    & 83.1    & 39.4     \\
Naive AN~\cite{kundu2020exploiting}     & NeurIPS 2020 &                        &                                    & 65.1    & 86.5    & 40.8     \\ \hline
TaI-DPT~\cite{guo2023texts}              & CVPR 2023    & Image Caption          & \multirow{2}{*}{Unlabeled}         & 65.1    & 88.3    & 46.5     \\
Ours                                                            & -            & ChatGLM                &                                    & 66.8    & 88.7    & 47.0   \\  \hline \hline
\end{tabular}
\label{tab:sota}
\end{table*}

\subsection{Parameter Analysis}
\begin{figure}[t]
    \centering
    % \begin{tabular}{cc}
    %     \includegraphics[width=0.9\linewidth]{figs/prompt.pdf} \\  \includegraphics[width=0.9\linewidth]{figs/weight.pdf}
    % \end{tabular}
    % \begin{tabular}{cc}
    %     \includegraphics[width=0.47\linewidth]{figs/prompt.pdf} &  \includegraphics[width=0.47\linewidth]{figs/weight.pdf}
    % \end{tabular}
    \includegraphics[width=0.47\textwidth]{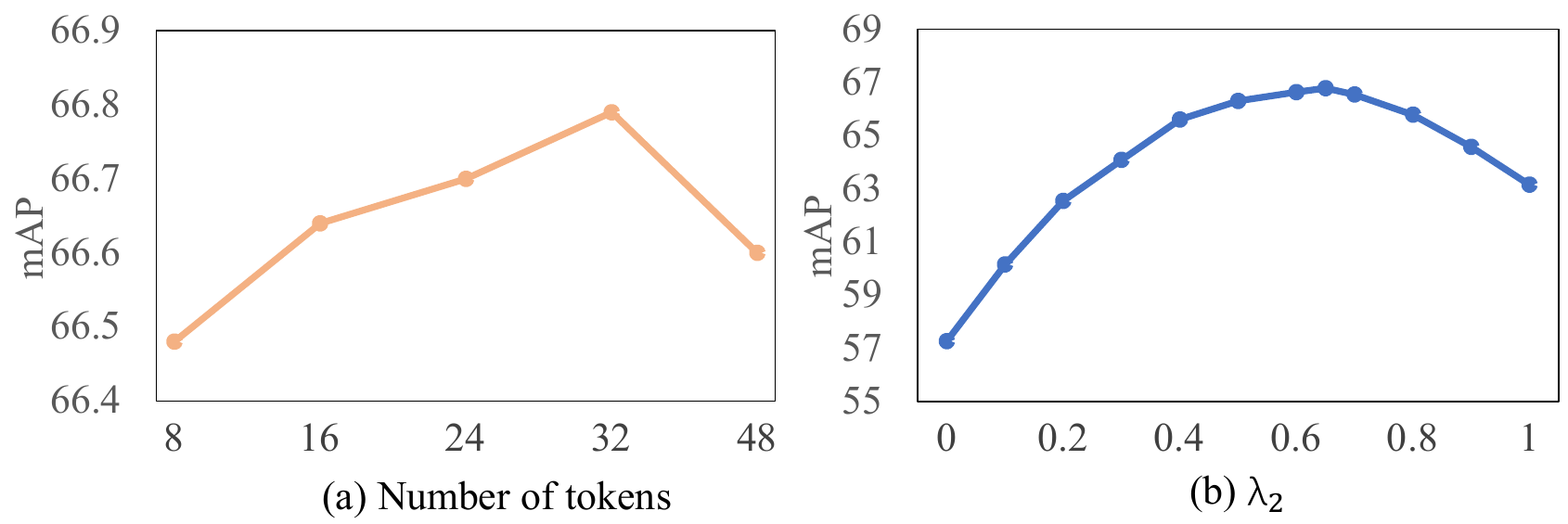}
    
    \caption{Results on MS-COCO. (a) Analysis of the effect of  number of tokens. (b) Analysis of the effect of weight between global and local prompts, \textit{i.e.} $\lambda_2$ in Eq.(\ref{eq:inference}).}
    \label{fig:parameter}
\end{figure}

% \noindent \textbf{Number of different types of token.}
% Our hierarchical prompts contain three types of tokens: shared tokens, partial-shared tokens and category-specific tokens. And we further split the partial-shared tokens into two levels: tokens of level \#1  are shared among more categories compared to those of level \#2.
% In Table~\ref{tab:abl_mix_tokens}, we present the results corresponding to various combinations of token quantities within our hierarchical prompts. 
% The similar mAP scores indicate that our method is robust to the numbers of different token types.
\noindent \textbf{Number of different types of token.}
The performances of different numbers of different token types in  hierarchical prompts on  MS-COCO are shown in Figure~\ref{fig:parameter} (a). We observe that the performance increases as the token number increases,  but a number larger than 32 brings negative impact. Moreover, we also analyze the composition of different types of tokens in Table~\ref{tab:abl_mix_tokens}. Note that the partial-shared tokens are split into two parts: ``P-S \#1" are coarse parts that tokens are shared over more categories than that of ``P-S \#2". We observe similar mAP scores for different configurations, indicating that our method is robust to the numbers of different token types.

\noindent \textbf{Weights of global and local prompts.}
To evaluate the contributions of the global and local prompts to the final performance, we conducted experiments with varied weights assigned to each branch, as depicted in Figure~\ref{fig:parameter} (b).
We observe that as the weight allocated to the global prompts increases, the performance initially rises, peaking at a weight of 0.65, then gradually declines. This trend demonstrates that the global branch plays a more critical role, but the local branch is also necessary.

% To evaluate how much the global branch and local branch contribute to the final performance, we set different weights for them and the results are shown in Figure~\ref{fig:parameter} (b).
% We observe that when the weight of global prompts increases, the performance first increases and reaches the maximum when the weight is 0.65 and then gradually decreases, demonstrating that global prompts play a more important role but the local prompts are also necessary.
% \begin{table}[t]
% \centering
% \scalebox{1}{\begin{tabular}{c|c}
% \hline
% \multicolumn{1}{c|}{prompt length} & \multicolumn{1}{|c}{mAP} \\ \hline
% \multicolumn{1}{c|}{8}    & 66.48  \\
% \multicolumn{1}{c|}{16}    & 66.64  \\
% \multicolumn{1}{c|}{24}    & 66.70  \\
% \multicolumn{1}{c|}{32}    & 66.79   \\
% \hline
% \end{tabular}}
% \end{table}	

% \noindent \textbf{global and local.}
% \begin{table}[h]
% \centering
% \scalebox{1}{\begin{tabular}{c|c}
% \hline
% \multicolumn{1}{c|}{global and local} & \multicolumn{1}{|c}{mAP} \\ \hline
% \multicolumn{1}{c|}{0.6}    & 66.64  \\
% \multicolumn{1}{c|}{0.65}    & 66.79  \\
% \multicolumn{1}{c|}{0.7}    & 66.55   \\
% \multicolumn{1}{c|}{0.75}    & 66.32   \\
% \hline
% \end{tabular}}
% \end{table}	

\begin{figure*}
    \centering
    \includegraphics[scale=0.38]{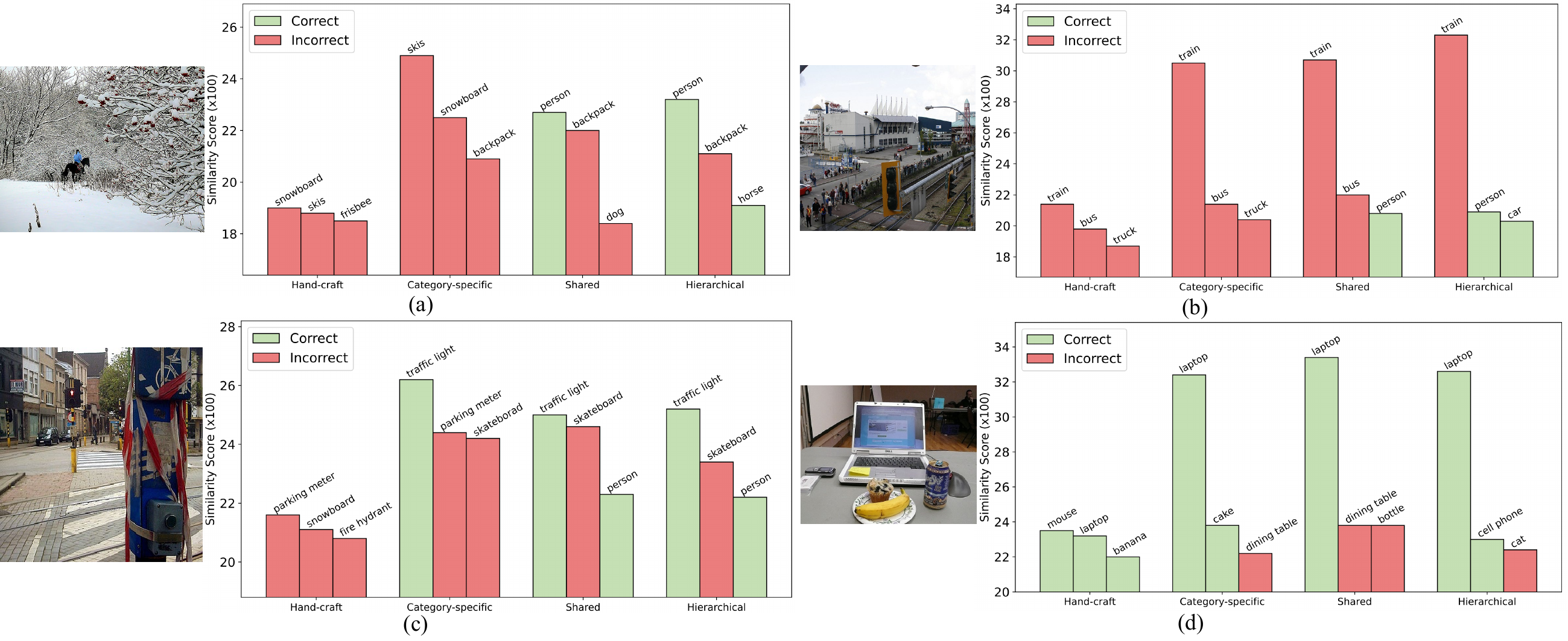}
    \caption{Visualization of top-3 predicated categories by different prompts.}
    \label{fig:vis}
\end{figure*}

\subsection{Comparison with State-of-the-art Methods}
We compare the proposed method with several state-of-the-art methods at different annotations levels of training images, including  fully labeled methods (SRN~\cite{zhu2017learning}, ML-GCN~\cite{chen2019multi}, and ASL~\cite{ridnik2021asymmetric}),  partially labeled methods (SARB~\cite{pu2022semantic}, SST~\cite{chen2022structured}, and DualCoOp~\cite{sun2022dualcoop}), one labeled methods (LL-R~\cite{kim2022large} and G$^2$NetPL ~\cite{abdelfattah2022g2netpl}), and unlabeled methods (LSAN~\cite{szegedy2016rethinking}, WAN~\cite{mac2019presence}, Curriculum~\cite{durand2019learning}, and Naive AN~\cite{kundu2020exploiting}). We also compare our method with a recent method TaI-DPT~\cite{guo2023texts} that uses image captions for training. 

Table~\ref{tab:sota} shows the comparison results on  VOC2007, MS-COCO, and NUS-WIDE. We have observations as follows: (1) Our method outperforms all the unsupervised methods using unlabeled training images, underscoring the superiority of comprehensive knowledge of objects stored in ChatGLM; (2) Our method exhibits a slight performance advantage over TaI-DPT, which is trained on human-written image captions. This result suggests that ChatGLM has the ability to emulate human understanding, further validating the effectiveness of our method;
%Our method demonstrates a slight performance edge over TaI-DPT, which trains on human-written image captions, affirming the LLM's capacity to emulate human understanding. 
%(3) A significant performance drop (more than 12\% on MS-COCO) between our method and the fully labeled methods demonstrates the persistence of modality gap of CLIP between language and image, despite the vast number of paired examples used in training.
(3) The performance of our method significantly drops compared to the fully labeled methods, probably due to the domain gap between the training data of CLIP and the target data of specific task. %Therefore, how to bridge the domain gap in CLIP using pre-trained models is a potential future direction.

\subsection{Qualitative Analysis}
%In Figure~\ref{fig:vis}, we present the top-3 category predictions made by models using different prompt methods. The model using hierarchical prompts shows enhanced accuracy in predicting object categories, as indicated by instances (a), (b), and (c). This improvement is especially pronounced when predicting smaller objects,  highlighting the effectiveness of hierarchical prompts over other types. However, in instance (d), the model utilizing hand-craft prompts predict a higher number of correct categories. This might be attributed to the meticulous design of the prompts, incorporating certain human prior knowledge. For example, in designing the prompt for the category "mouse," we use the term "computer mouse," which is more relevant to its contextual usage.
% Figure~\ref{fig:vis} shows the top-3 category predictions using different prompts. We observe that the hierarchical prompts achieve better performance, especially on smaller objects, as shown in (a), (b) and (c). However, as shown in (b), the top-1 wrong predicted category is ``train" due to overweigh the global image features, which looks like a train station. And as shown in (d), the hand-craft prompts perform best, probably due to that the meticulous design of hand-craft prompts integrates certain human prior knowledge. For example, in designing the prompt for the category ``mouse",  we use the term of ``computer mouse", which is more relevant to its contextual usage.
Figure~\ref{fig:vis} illustrates the top-3 category predictions of different prompts. Notably, the hierarchical prompts achieve better performance,  especially on smaller objects, as shown in (a), (b), and (c). However, as depicted in (b), an instance of incorrect top-1 prediction arises in the category labeled ``train", likely due to an excessive emphasis on global image features, resembling a train station. Conversely, as shown in (d), the hand-craft prompts demonstrate superior performance,  probably due to the meticulous design of hand-craft prompts integrating certain human prior knowledge.
For instance, when designing the prompt for the ``mouse" category, we use the term ``computer mouse", aligning more closely with its contextual usage to improve the performance.

\section{Conclusion}
%We have introduced an innovative data-free framework for multi-label image recognition, harnessing the expansive knowledge within Large Language Models (LLMs) to adapt Vision-Language pre-trained Models (VLMs) like CLIP through prompt tuning. 
%Our method involves querying the LLM with well-designed questions, facilitating the comprehensive extraction of object characteristics and their contextual relationships.
%We have proposed a hierarchical prompt learning method that considers label dependencies among object categories. This method enables the partial sharing of prompt tokens across categories, leveraging their similar attributes and co-occurrence patterns.
We have presented a novel data-free framework for multi-label image recognition, which leverages enriched text descriptions powered by LLMs such as ChatGLM to well adapt VLMs like CLIP through prompt tuning.  By first querying ChatGLM with well-designed questions and then learning hierarchical prompts with contextual relationships between  categories, our method successfully achieves promising results without any training data, which is evaluated by extensive experiments on three benchmark datasets. 
Our method provides an effective way to  explore the synergies between multiple pre-trained models for  visual recognition under data scarcity.
In the future, we are going to apply the proposed data-free framework to more computer vision tasks such as action recognition in videos.

% We have presented a novel data-free framework for multi-label image recognition,  which leverage the knowledge of Large Language Model (LLM) to adapt Vision-Language pre-trained Models (VLMs) like CLIP to multi-label classification by prompt tuning. 
% We introduce a meticulous process of querying the LLM with well-structured questions to gather comprehensive insights into object characteristics and their contextual relationships.
% Utilizing this knowledge, we proposed a hierarchical prompt learning method that considers label dependencies of object categories, allowing for the partial sharing of prompt tokens between different categories based on their similarities in attributes and co-occurrence patterns.
% Extensive experimental results on three benchmark datasets demonstrate the effectiveness of our method.
% Future research avenues could explore further advancements in leveraging textual descriptions for multi-label image recognition tasks, aiming to refine and expand the capabilities of data-free frameworks in the realm of computer vision

% \input{sec/1_intro}
% \input{sec/2_formatting}
% \input{sec/3_finalcopy}
{
    \small
    \bibliographystyle{ieeenat_fullname}
    \bibliography{main}

\begin{thebibliography}{64}
\providecommand{\natexlab}[1]{#1}
\providecommand{\url}[1]{\texttt{#1}}
\expandafter\ifx\csname urlstyle\endcsname\relax
  \providecommand{\doi}[1]{doi: #1}\else
  \providecommand{\doi}{doi: \begingroup \urlstyle{rm}\Url}\fi

\bibitem[Abdelfattah et~al.(2022{\natexlab{a}})Abdelfattah, Zhang, Fouda, Wang, and Wang]{abdelfattah2022g2netpl}
Rabab Abdelfattah, Xin Zhang, Mostafa~M Fouda, Xiaofeng Wang, and Song Wang.
\newblock G2netpl: Generic game-theoretic network for partial-label image classification.
\newblock \emph{BMVC}, 2022{\natexlab{a}}.

\bibitem[Abdelfattah et~al.(2022{\natexlab{b}})Abdelfattah, Zhang, Wu, Wu, Wang, and Wang]{abdelfattah2022plmcl}
Rabab Abdelfattah, Xin Zhang, Zhenyao Wu, Xinyi Wu, Xiaofeng Wang, and Song Wang.
\newblock Plmcl: Partial-label momentum curriculum learning for multi-label image classification.
\newblock In \emph{ECCV}, pages 39--55. Springer, 2022{\natexlab{b}}.

\bibitem[Abdelfattah et~al.(2023)Abdelfattah, Guo, Li, Wang, and Wang]{abdelfattah2023cdul}
Rabab Abdelfattah, Qing Guo, Xiaoguang Li, Xiaofeng Wang, and Song Wang.
\newblock Cdul: Clip-driven unsupervised learning for multi-label image classification.
\newblock In \emph{ICCV}, pages 1348--1357, 2023.

\bibitem[Alfassy et~al.(2019)Alfassy, Karlinsky, Aides, Shtok, Harary, Feris, Giryes, and Bronstein]{alfassy2019laso}
Amit Alfassy, Leonid Karlinsky, Amit Aides, Joseph Shtok, Sivan Harary, Rogerio Feris, Raja Giryes, and Alex~M Bronstein.
\newblock Laso: Label-set operations networks for multi-label few-shot learning.
\newblock In \emph{CVPR}, pages 6548--6557, 2019.

\bibitem[Ben-Cohen et~al.(2021)Ben-Cohen, Zamir, Ben-Baruch, Friedman, and Zelnik-Manor]{ben2021semantic}
Avi Ben-Cohen, Nadav Zamir, Emanuel Ben-Baruch, Itamar Friedman, and Lihi Zelnik-Manor.
\newblock Semantic diversity learning for zero-shot multi-label classification.
\newblock In \emph{ICCV}, pages 640--650, 2021.

\bibitem[Brown et~al.(2020)Brown, Mann, Ryder, Subbiah, Kaplan, Dhariwal, Neelakantan, Shyam, Sastry, Askell, et~al.]{brown2020language}
Tom Brown, Benjamin Mann, Nick Ryder, Melanie Subbiah, Jared~D Kaplan, Prafulla Dhariwal, Arvind Neelakantan, Pranav Shyam, Girish Sastry, Amanda Askell, et~al.
\newblock Language models are few-shot learners.
\newblock \emph{NeurIPS}, 33:\penalty0 1877--1901, 2020.

\bibitem[Bubeck et~al.(2023)Bubeck, Chandrasekaran, Eldan, Gehrke, Horvitz, Kamar, Lee, Lee, Li, Lundberg, et~al.]{bubeck2023sparks}
S{\'e}bastien Bubeck, Varun Chandrasekaran, Ronen Eldan, Johannes Gehrke, Eric Horvitz, Ece Kamar, Peter Lee, Yin~Tat Lee, Yuanzhi Li, Scott Lundberg, et~al.
\newblock Sparks of artificial general intelligence: Early experiments with gpt-4.
\newblock \emph{arXiv preprint arXiv:2303.12712}, 2023.

\bibitem[Cai et~al.(2023)Cai, Mao, Wu, Wang, Liang, Ge, Wu, You, Song, Xia, et~al.]{cai2023low}
Yuzhe Cai, Shaoguang Mao, Wenshan Wu, Zehua Wang, Yaobo Liang, Tao Ge, Chenfei Wu, Wang You, Ting Song, Yan Xia, et~al.
\newblock Low-code llm: Visual programming over llms.
\newblock \emph{arXiv preprint arXiv:2304.08103}, 2023.

\bibitem[Chen et~al.(2023)Chen, Liu, Wang, Zhang, Torr, Zhang, and Tang]{chen2023tem}
Guangyi Chen, Xiao Liu, Guangrun Wang, Kun Zhang, Philip~HS Torr, Xiao-Ping Zhang, and Yansong Tang.
\newblock Tem-adapter: Adapting image-text pretraining for video question answer.
\newblock In \emph{Proceedings of the IEEE/CVF International Conference on Computer Vision}, pages 13945--13955, 2023.

\bibitem[Chen et~al.(2019{\natexlab{a}})Chen, Xu, Hui, Wu, and Lin]{chen2019learning}
Tianshui Chen, Muxin Xu, Xiaolu Hui, Hefeng Wu, and Liang Lin.
\newblock Learning semantic-specific graph representation for multi-label image recognition.
\newblock In \emph{ICCV}, pages 522--531, 2019{\natexlab{a}}.

\bibitem[Chen et~al.(2022)Chen, Pu, Wu, Xie, and Lin]{chen2022structured}
Tianshui Chen, Tao Pu, Hefeng Wu, Yuan Xie, and Liang Lin.
\newblock Structured semantic transfer for multi-label recognition with partial labels.
\newblock In \emph{AAAI}, pages 339--346, 2022.

\bibitem[Chen et~al.(2019{\natexlab{b}})Chen, Wei, Wang, and Guo]{chen2019multi}
Zhao-Min Chen, Xiu-Shen Wei, Peng Wang, and Yanwen Guo.
\newblock Multi-label image recognition with graph convolutional networks.
\newblock In \emph{CVPR}, pages 5177--5186, 2019{\natexlab{b}}.

\bibitem[Chua et~al.(2009)Chua, Tang, Hong, Li, Luo, and Zheng]{chua2009nus}
Tat-Seng Chua, Jinhui Tang, Richang Hong, Haojie Li, Zhiping Luo, and Yantao Zheng.
\newblock Nus-wide: a real-world web image database from national university of singapore.
\newblock In \emph{Proceedings of the ACM international conference on image and video retrieval}, pages 1--9, 2009.

\bibitem[Du et~al.(2022)Du, Qian, Liu, Ding, Qiu, Yang, and Tang]{du2022glm}
Zhengxiao Du, Yujie Qian, Xiao Liu, Ming Ding, Jiezhong Qiu, Zhilin Yang, and Jie Tang.
\newblock Glm: General language model pretraining with autoregressive blank infilling.
\newblock In \emph{ACL}, pages 320--335, 2022.

\bibitem[Durand et~al.(2019)Durand, Mehrasa, and Mori]{durand2019learning}
Thibaut Durand, Nazanin Mehrasa, and Greg Mori.
\newblock Learning a deep convnet for multi-label classification with partial labels.
\newblock In \emph{CVPR}, pages 647--657, 2019.

\bibitem[Everingham et~al.(2010)Everingham, Van~Gool, Williams, Winn, and Zisserman]{everingham2010pascal}
Mark Everingham, Luc Van~Gool, Christopher~KI Williams, John Winn, and Andrew Zisserman.
\newblock The pascal visual object classes (voc) challenge.
\newblock \emph{IJCV}, 88:\penalty0 303--338, 2010.

\bibitem[Gao and Zhou(2020)]{gao2020multi}
Bin-Bin Gao and Hong-Yu Zhou.
\newblock Multi-label image recognition with multi-class attentional regions.
\newblock \emph{arXiv e-prints}, pages arXiv--2007, 2020.

\bibitem[Gao et~al.(2023)Gao, Geng, Zhang, Ma, Fang, Zhang, Li, and Qiao]{gao2023clip}
Peng Gao, Shijie Geng, Renrui Zhang, Teli Ma, Rongyao Fang, Yongfeng Zhang, Hongsheng Li, and Yu Qiao.
\newblock Clip-adapter: Better vision-language models with feature adapters.
\newblock \emph{IJCV}, pages 1--15, 2023.

\bibitem[Gong et~al.(2013)Gong, Jia, Leung, Toshev, and Ioffe]{gong2013deep}
Yunchao Gong, Yangqing Jia, Thomas Leung, Alexander Toshev, and Sergey Ioffe.
\newblock Deep convolutional ranking for multilabel image annotation.
\newblock \emph{arXiv preprint arXiv:1312.4894}, 2013.

\bibitem[Guo et~al.(2023{\natexlab{a}})Guo, Dong, Ji, Bai, Guo, and Zuo]{guo2023texts}
Zixian Guo, Bowen Dong, Zhilong Ji, Jinfeng Bai, Yiwen Guo, and Wangmeng Zuo.
\newblock Texts as images in prompt tuning for multi-label image recognition.
\newblock In \emph{CVPR}, pages 2808--2817, 2023{\natexlab{a}}.

\bibitem[Guo et~al.(2023{\natexlab{b}})Guo, Zhang, Qiu, Ma, Miao, He, and Cui]{guo2023calip}
Ziyu Guo, Renrui Zhang, Longtian Qiu, Xianzheng Ma, Xupeng Miao, Xuming He, and Bin Cui.
\newblock Calip: Zero-shot enhancement of clip with parameter-free attention.
\newblock In \emph{AAAI}, pages 746--754, 2023{\natexlab{b}}.

\bibitem[Gupta and Kembhavi(2023)]{gupta2023visual}
Tanmay Gupta and Aniruddha Kembhavi.
\newblock Visual programming: Compositional visual reasoning without training.
\newblock In \emph{CVPR}, pages 14953--14962, 2023.

\bibitem[He et~al.(2018)He, Xu, Guo, Xu, and Tao]{he2018reinforced}
Shiyi He, Chang Xu, Tianyu Guo, Chao Xu, and Dacheng Tao.
\newblock Reinforced multi-label image classification by exploring curriculum.
\newblock In \emph{AAAI}, 2018.

\bibitem[Huynh and Elhamifar(2020)]{huynh2020shared}
Dat Huynh and Ehsan Elhamifar.
\newblock A shared multi-attention framework for multi-label zero-shot learning.
\newblock In \emph{CVPR}, pages 8776--8786, 2020.

\bibitem[Ji et~al.(2020)Ji, Cui, Li, Jiang, Xiang, Hospedales, and Fu]{ji2020deep}
Zhong Ji, Biying Cui, Huihui Li, Yu-Gang Jiang, Tao Xiang, Timothy Hospedales, and Yanwei Fu.
\newblock Deep ranking for image zero-shot multi-label classification.
\newblock \emph{IEEE TIP}, 29:\penalty0 6549--6560, 2020.

\bibitem[Jia et~al.(2022)Jia, Tang, Chen, Cardie, Belongie, Hariharan, and Lim]{jia2022visual}
Menglin Jia, Luming Tang, Bor-Chun Chen, Claire Cardie, Serge Belongie, Bharath Hariharan, and Ser-Nam Lim.
\newblock Visual prompt tuning.
\newblock In \emph{ECCV}, pages 709--727. Springer, 2022.

\bibitem[Kim et~al.(2022)Kim, Kim, Akata, and Lee]{kim2022large}
Youngwook Kim, Jae~Myung Kim, Zeynep Akata, and Jungwoo Lee.
\newblock Large loss matters in weakly supervised multi-label classification.
\newblock In \emph{CVPR}, pages 14156--14165, 2022.

\bibitem[Kundu and Tighe(2020)]{kundu2020exploiting}
Kaustav Kundu and Joseph Tighe.
\newblock Exploiting weakly supervised visual patterns to learn from partial annotations.
\newblock \emph{NeurIPS}, 33:\penalty0 561--572, 2020.

\bibitem[Lee et~al.(2018)Lee, Fang, Yeh, and Wang]{lee2018multi}
Chung-Wei Lee, Wei Fang, Chih-Kuan Yeh, and Yu-Chiang~Frank Wang.
\newblock Multi-label zero-shot learning with structured knowledge graphs.
\newblock In \emph{CVPR}, pages 1576--1585, 2018.

\bibitem[Li et~al.(2017)Li, Song, and Luo]{li2017improving}
Yuncheng Li, Yale Song, and Jiebo Luo.
\newblock Improving pairwise ranking for multi-label image classification.
\newblock In \emph{CVPR}, pages 3617--3625, 2017.

\bibitem[Lin et~al.(2014)Lin, Maire, Belongie, Hays, Perona, Ramanan, Doll{\'a}r, and Zitnick]{lin2014microsoft}
Tsung-Yi Lin, Michael Maire, Serge Belongie, James Hays, Pietro Perona, Deva Ramanan, Piotr Doll{\'a}r, and C~Lawrence Zitnick.
\newblock Microsoft coco: Common objects in context.
\newblock In \emph{ECCV}, pages 740--755. Springer, 2014.

\bibitem[Liu et~al.(2017)Liu, Xiang, Hospedales, Yang, and Sun]{liu2017semantic}
Feng Liu, Tao Xiang, Timothy~M Hospedales, Wankou Yang, and Changyin Sun.
\newblock Semantic regularisation for recurrent image annotation.
\newblock In \emph{CVPR}, pages 2872--2880, 2017.

\bibitem[Liu and Tsang(2015)]{liu2015optimality}
Weiwei Liu and Ivor Tsang.
\newblock On the optimality of classifier chain for multi-label classification.
\newblock \emph{NeurIPS}, 28, 2015.

\bibitem[Mac~Aodha et~al.(2019)Mac~Aodha, Cole, and Perona]{mac2019presence}
Oisin Mac~Aodha, Elijah Cole, and Pietro Perona.
\newblock Presence-only geographical priors for fine-grained image classification.
\newblock In \emph{ICCV}, pages 9596--9606, 2019.

\bibitem[Misra et~al.(2016)Misra, Lawrence~Zitnick, Mitchell, and Girshick]{misra2016seeing}
Ishan Misra, C Lawrence~Zitnick, Margaret Mitchell, and Ross Girshick.
\newblock Seeing through the human reporting bias: Visual classifiers from noisy human-centric labels.
\newblock In \emph{CVPR}, pages 2930--2939, 2016.

\bibitem[Pu et~al.(2022)Pu, Chen, Wu, and Lin]{pu2022semantic}
Tao Pu, Tianshui Chen, Hefeng Wu, and Liang Lin.
\newblock Semantic-aware representation blending for multi-label image recognition with partial labels.
\newblock In \emph{AAAI}, pages 2091--2098, 2022.

\bibitem[Radford et~al.(2021)Radford, Kim, Hallacy, Ramesh, Goh, Agarwal, Sastry, Askell, Mishkin, Clark, et~al.]{radford2021learning}
Alec Radford, Jong~Wook Kim, Chris Hallacy, Aditya Ramesh, Gabriel Goh, Sandhini Agarwal, Girish Sastry, Amanda Askell, Pamela Mishkin, Jack Clark, et~al.
\newblock Learning transferable visual models from natural language supervision.
\newblock In \emph{International conference on machine learning}, pages 8748--8763. PMLR, 2021.

\bibitem[Ridnik et~al.(2021)Ridnik, Ben-Baruch, Zamir, Noy, Friedman, Protter, and Zelnik-Manor]{ridnik2021asymmetric}
Tal Ridnik, Emanuel Ben-Baruch, Nadav Zamir, Asaf Noy, Itamar Friedman, Matan Protter, and Lihi Zelnik-Manor.
\newblock Asymmetric loss for multi-label classification.
\newblock In \emph{ICCV}, pages 82--91, 2021.

\bibitem[Sarafianos et~al.(2018)Sarafianos, Xu, and Kakadiaris]{sarafianos2018deep}
Nikolaos Sarafianos, Xiang Xu, and Ioannis~A Kakadiaris.
\newblock Deep imbalanced attribute classification using visual attention aggregation.
\newblock In \emph{ECCV}, pages 680--697, 2018.

\bibitem[Simon et~al.(2022)Simon, Koniusz, and Harandi]{simon2022meta}
Christian Simon, Piotr Koniusz, and Mehrtash Harandi.
\newblock Meta-learning for multi-label few-shot classification.
\newblock In \emph{WACV}, pages 3951--3960, 2022.

\bibitem[Singha et~al.(2023)Singha, Pal, Jha, and Banerjee]{singha2023ad}
Mainak Singha, Harsh Pal, Ankit Jha, and Biplab Banerjee.
\newblock Ad-clip: Adapting domains in prompt space using clip.
\newblock In \emph{ICCV}, pages 4355--4364, 2023.

\bibitem[Sohn et~al.(2023)Sohn, Chang, Lezama, Polania, Zhang, Hao, Essa, and Jiang]{sohn2023visual}
Kihyuk Sohn, Huiwen Chang, Jos{\'e} Lezama, Luisa Polania, Han Zhang, Yuan Hao, Irfan Essa, and Lu Jiang.
\newblock Visual prompt tuning for generative transfer learning.
\newblock In \emph{CVPR}, pages 19840--19851, 2023.

\bibitem[Sun et~al.(2022)Sun, Hu, and Saenko]{sun2022dualcoop}
Ximeng Sun, Ping Hu, and Kate Saenko.
\newblock Dualcoop: Fast adaptation to multi-label recognition with limited annotations.
\newblock \emph{NeurIPS}, 35:\penalty0 30569--30582, 2022.

\bibitem[Sung et~al.(2022)Sung, Cho, and Bansal]{sung2022vl}
Yi-Lin Sung, Jaemin Cho, and Mohit Bansal.
\newblock Vl-adapter: Parameter-efficient transfer learning for vision-and-language tasks.
\newblock In \emph{CVPR}, pages 5227--5237, 2022.

\bibitem[Szegedy et~al.(2016)Szegedy, Vanhoucke, Ioffe, Shlens, and Wojna]{szegedy2016rethinking}
Christian Szegedy, Vincent Vanhoucke, Sergey Ioffe, Jon Shlens, and Zbigniew Wojna.
\newblock Rethinking the inception architecture for computer vision.
\newblock In \emph{CVPR}, pages 2818--2826, 2016.

\bibitem[Upadhyay et~al.(2023)Upadhyay, Karthik, Mancini, and Akata]{upadhyay2023probvlm}
Uddeshya Upadhyay, Shyamgopal Karthik, Massimiliano Mancini, and Zeynep Akata.
\newblock Probvlm: Probabilistic adapter for frozen vison-language models.
\newblock In \emph{ICCV}, pages 1899--1910, 2023.

\bibitem[Wang et~al.(2016)Wang, Yang, Mao, Huang, Huang, and Xu]{wang2016cnn}
Jiang Wang, Yi Yang, Junhua Mao, Zhiheng Huang, Chang Huang, and Wei Xu.
\newblock Cnn-rnn: A unified framework for multi-label image classification.
\newblock In \emph{CVPR}, pages 2285--2294, 2016.

\bibitem[Wang et~al.(2020)Wang, He, Li, Long, Zhou, Ma, and Wen]{wang2020multi}
Ya Wang, Dongliang He, Fu Li, Xiang Long, Zhichao Zhou, Jinwen Ma, and Shilei Wen.
\newblock Multi-label classification with label graph superimposing.
\newblock In \emph{AAAI}, pages 12265--12272, 2020.

\bibitem[Wang et~al.(2017)Wang, Chen, Li, Xu, and Lin]{wang2017multi}
Zhouxia Wang, Tianshui Chen, Guanbin Li, Ruijia Xu, and Liang Lin.
\newblock Multi-label image recognition by recurrently discovering attentional regions.
\newblock In \emph{ICCV}, pages 464--472, 2017.

\bibitem[Wei et~al.(2022)Wei, Tay, Bommasani, Raffel, Zoph, Borgeaud, Yogatama, Bosma, Zhou, Metzler, Chi, Hashimoto, Vinyals, Liang, Dean, and Fedus]{wei2022emergent}
Jason Wei, Yi Tay, Rishi Bommasani, Colin Raffel, Barret Zoph, Sebastian Borgeaud, Dani Yogatama, Maarten Bosma, Denny Zhou, Donald Metzler, Ed~H. Chi, Tatsunori Hashimoto, Oriol Vinyals, Percy Liang, Jeff Dean, and William Fedus.
\newblock Emergent abilities of large language models.
\newblock \emph{Transactions on Machine Learning Research}, 2022.
\newblock Survey Certification.

\bibitem[Wei et~al.(2015)Wei, Xia, Lin, Huang, Ni, Dong, Zhao, and Yan]{wei2015hcp}
Yunchao Wei, Wei Xia, Min Lin, Junshi Huang, Bingbing Ni, Jian Dong, Yao Zhao, and Shuicheng Yan.
\newblock Hcp: A flexible cnn framework for multi-label image classification.
\newblock \emph{IEEE TPAMI}, 38\penalty0 (9):\penalty0 1901--1907, 2015.

\bibitem[Xu et~al.(2023)Xu, Zhang, Wei, Hu, and Bai]{xu2023side}
Mengde Xu, Zheng Zhang, Fangyun Wei, Han Hu, and Xiang Bai.
\newblock Side adapter network for open-vocabulary semantic segmentation.
\newblock In \emph{CVPR}, pages 2945--2954, 2023.

\bibitem[Yang et~al.(2023)Yang, Chen, Qian, Madaan, Iyengar, Fouhey, and Chai]{yang2023llm}
Jianing Yang, Xuweiyi Chen, Shengyi Qian, Nikhil Madaan, Madhavan Iyengar, David~F Fouhey, and Joyce Chai.
\newblock Llm-grounder: Open-vocabulary 3d visual grounding with large language model as an agent.
\newblock \emph{arXiv preprint arXiv:2309.12311}, 2023.

\bibitem[Yang et~al.(2022)Yang, Gan, Wang, Hu, Lu, Liu, and Wang]{yang2022empirical}
Zhengyuan Yang, Zhe Gan, Jianfeng Wang, Xiaowei Hu, Yumao Lu, Zicheng Liu, and Lijuan Wang.
\newblock An empirical study of gpt-3 for few-shot knowledge-based vqa.
\newblock In \emph{AAAI}, pages 3081--3089, 2022.

\bibitem[Yao et~al.(2021)Yao, Zhang, Zhang, Liu, Chua, and Sun]{yao2021cpt}
Yuan Yao, Ao Zhang, Zhengyan Zhang, Zhiyuan Liu, Tat-Seng Chua, and Maosong Sun.
\newblock Cpt: Colorful prompt tuning for pre-trained vision-language models.
\newblock \emph{arXiv preprint arXiv:2109.11797}, 2021.

\bibitem[Yazici et~al.(2020)Yazici, Gonzalez-Garcia, Ramisa, Twardowski, and Weijer]{yazici2020orderless}
Vacit~Oguz Yazici, Abel Gonzalez-Garcia, Arnau Ramisa, Bartlomiej Twardowski, and Joost van~de Weijer.
\newblock Orderless recurrent models for multi-label classification.
\newblock In \emph{CVPR}, pages 13440--13449, 2020.

\bibitem[Ye et~al.(2020)Ye, He, Peng, Wu, and Qiao]{ye2020attention}
Jin Ye, Junjun He, Xiaojiang Peng, Wenhao Wu, and Yu Qiao.
\newblock Attention-driven dynamic graph convolutional network for multi-label image recognition.
\newblock In \emph{ECCV}, pages 649--665. Springer, 2020.

\bibitem[Zhang et~al.(2018)Zhang, Wu, Shen, Zhang, and Lu]{zhang2018multilabel}
Junjie Zhang, Qi Wu, Chunhua Shen, Jian Zhang, and Jianfeng Lu.
\newblock Multilabel image classification with regional latent semantic dependencies.
\newblock \emph{IEEE TMM}, 20\penalty0 (10):\penalty0 2801--2813, 2018.

\bibitem[Zhang et~al.(2022)Zhang, Zhang, Fang, Gao, Li, Dai, Qiao, and Li]{zhang2022tip}
Renrui Zhang, Wei Zhang, Rongyao Fang, Peng Gao, Kunchang Li, Jifeng Dai, Yu Qiao, and Hongsheng Li.
\newblock Tip-adapter: Training-free adaption of clip for few-shot classification.
\newblock In \emph{ECCV}, pages 493--510. Springer, 2022.

\bibitem[Zhang et~al.(2023)Zhang, Liu, Zeng, Ooi, Tang, and Zhuang]{zhang2023learning}
Wenqiao Zhang, Changshuo Liu, Lingze Zeng, Bengchin Ooi, Siliang Tang, and Yueting Zhuang.
\newblock Learning in imperfect environment: Multi-label classification with long-tailed distribution and partial labels.
\newblock In \emph{ICCV}, pages 1423--1432, 2023.

\bibitem[Zhou et~al.(2022{\natexlab{a}})Zhou, Yang, Loy, and Liu]{zhou2022conditional}
Kaiyang Zhou, Jingkang Yang, Chen~Change Loy, and Ziwei Liu.
\newblock Conditional prompt learning for vision-language models.
\newblock In \emph{CVPR}, pages 16816--16825, 2022{\natexlab{a}}.

\bibitem[Zhou et~al.(2022{\natexlab{b}})Zhou, Yang, Loy, and Liu]{zhou2022learning}
Kaiyang Zhou, Jingkang Yang, Chen~Change Loy, and Ziwei Liu.
\newblock Learning to prompt for vision-language models.
\newblock \emph{IJCV}, 130\penalty0 (9):\penalty0 2337--2348, 2022{\natexlab{b}}.

\bibitem[Zhu et~al.(2023)Zhu, Niu, Han, Wu, and Zhang]{zhu2023prompt}
Beier Zhu, Yulei Niu, Yucheng Han, Yue Wu, and Hanwang Zhang.
\newblock Prompt-aligned gradient for prompt tuning.
\newblock In \emph{ICCV}, pages 15659--15669, 2023.

\bibitem[Zhu et~al.(2017)Zhu, Li, Ouyang, Yu, and Wang]{zhu2017learning}
Feng Zhu, Hongsheng Li, Wanli Ouyang, Nenghai Yu, and Xiaogang Wang.
\newblock Learning spatial regularization with image-level supervisions for multi-label image classification.
\newblock In \emph{CVPR}, pages 5513--5522, 2017.

\end{thebibliography}
}


\end{document}